\documentclass[10pt,journal,compsoc]{IEEEtran}
\usepackage[utf8]{inputenc}
\usepackage{tikz}
\usepackage{comment}
\usepackage{amsmath,amssymb} 
\usepackage{color}
\usepackage{wrapfig}
\usepackage{rotating}
\usepackage{epsfig}
\usepackage{graphicx}
\usepackage{multirow}
\usepackage{subcaption}
\usepackage{tabularx}
\usepackage{floatrow}
\usepackage{booktabs}
\usepackage{cuted}
\usepackage{multirow}
\usepackage{capt-of,etoolbox}	
\usepackage{multirow}
\usepackage{enumitem}
\usepackage[normalem]{ulem}
\usepackage{pifont}

\usepackage[pagebackref,breaklinks,colorlinks]{hyperref}
\newcommand{\name}{{Inst-Inpaint}}
\newcommand{\dataset}{{GQA-Inpaint}}

\ifCLASSOPTIONcompsoc
  \usepackage[nocompress]{cite}
\else
  \usepackage{cite}
\fi

\begin{document}

\title{Inst-Inpaint: Instructing to Remove Objects with Diffusion Models}

\author{
Ahmet Burak Yildirim,  Vedat Baday,  Erkut Erdem,  Aykut Erdem, and  Aysegul Dundar\
\IEEEcompsocitemizethanks{
\IEEEcompsocthanksitem A. B. Yildirim and A. Dundar  are with the Department of Computer Science, Bilkent  University,
Ankara, Turkey, {a.yildirim@bilkent.edu.tr} and {adundar@cs.bilkent.edu.tr}%
\IEEEcompsocthanksitem V. Baday and E. Erdem are with the Department of Computer Science, Hacettepe  University,
Ankara, Turkey.%
\IEEEcompsocthanksitem A. Erdem is with the Department of Computer Science, Ko\c{c} University,
Istanbul, Turkey.%
}
}


\IEEEtitleabstractindextext{
\begin{abstract}

Image inpainting task refers to erasing unwanted pixels from images and filling them in a semantically consistent and realistic way.
Traditionally, the pixels that are wished to be erased are defined with binary masks.
From the application point of view, a user needs to generate the masks for the objects they would like to remove which can be time-consuming and prone to errors.
In this work, we are interested in an image inpainting algorithm that estimates which object to be removed based on natural language input and removes it, simultaneously.
For this purpose, first, we construct a dataset named \textbf{\dataset} for this task. 
Second, we present a novel inpainting framework, \textbf{\name}, that can remove objects from images based on the instructions given as text prompts.
We set various GAN and diffusion-based baselines and run experiments on synthetic and real image datasets. 
We compare methods with different evaluation metrics that measure the quality and accuracy of the models and show significant quantitative and qualitative improvements. 
 Please refer to the web page of the project for the released code and dataset: \href{http://instinpaint.abyildirim.com/}{http://instinpaint.abyildirim.com/}
\end{abstract}

}

\maketitle

\IEEEdisplaynontitleabstractindextext
\IEEEpeerreviewmaketitle


\section{Introduction}
Image inpainting refers to the task of removing unwanted objects and/or filling in missing regions within an image~\cite{barnes2009patchmatch, pathak2016context, liu2018image}. It is considered highly challenging as the inpainted regions in the resulting images should be in harmony with the rest of the images. Inpainting approaches should take into account the global image context while filling in the missing pixels. Traditionally, existing studies use binary masks to highlight the missing regions. As in many image enhancement tasks, the recent image inpainting approaches utilize deep-learning-based frameworks equipped with generative priors and formulate the inpainting task as a supervised learning problem  \cite{yu2019free, liu2022partial, yu2022high}. That is, training data includes pairs of input and erased images acting as source and target data, respectively, and models are generally trained using a reconstruction-based objective. Although we have seen continuous improvements in the performance of inpainting approaches over the past years, these methods still require user-specified masks provided at inference time. These masks can be provided in terms of thin or thick brush strokes or squares, but drawing these masks is often tedious and prone to errors.


Recently, text-based image generation and editing have gained a lot of attention \cite{rombach2022high,bar2022text2live}). Especially when trained on large-scale image-text data, models like DALL$\cdot$E2~\cite{ramesh2022hierarchical}, Stable Diffusion~\cite{rombach2021highresolution}, Imagen~\cite{saharia2022photorealistic} and Parti~\cite{yu2022scaling} show exceptional generative capabilities while effectively capturing the compositional structure of images. Moreover, using text enables a more natural and user-friendly control over the images. To our interest, the researchers also demonstrated that these approaches could be easily adapted to image inpainting by additionally conditioning the models on binary masks, and telling what to fill in inside the mask. That said, that approach is better suited for playing with the attributes of an object or adding a new scene element, but may not work that well for removing objects. For example, please see the Blended Latent Diffusion model~\cite{avrahami2022blended}.

\begin{figure*}[!t]
    \centering
    \includegraphics[width=0.99\linewidth]{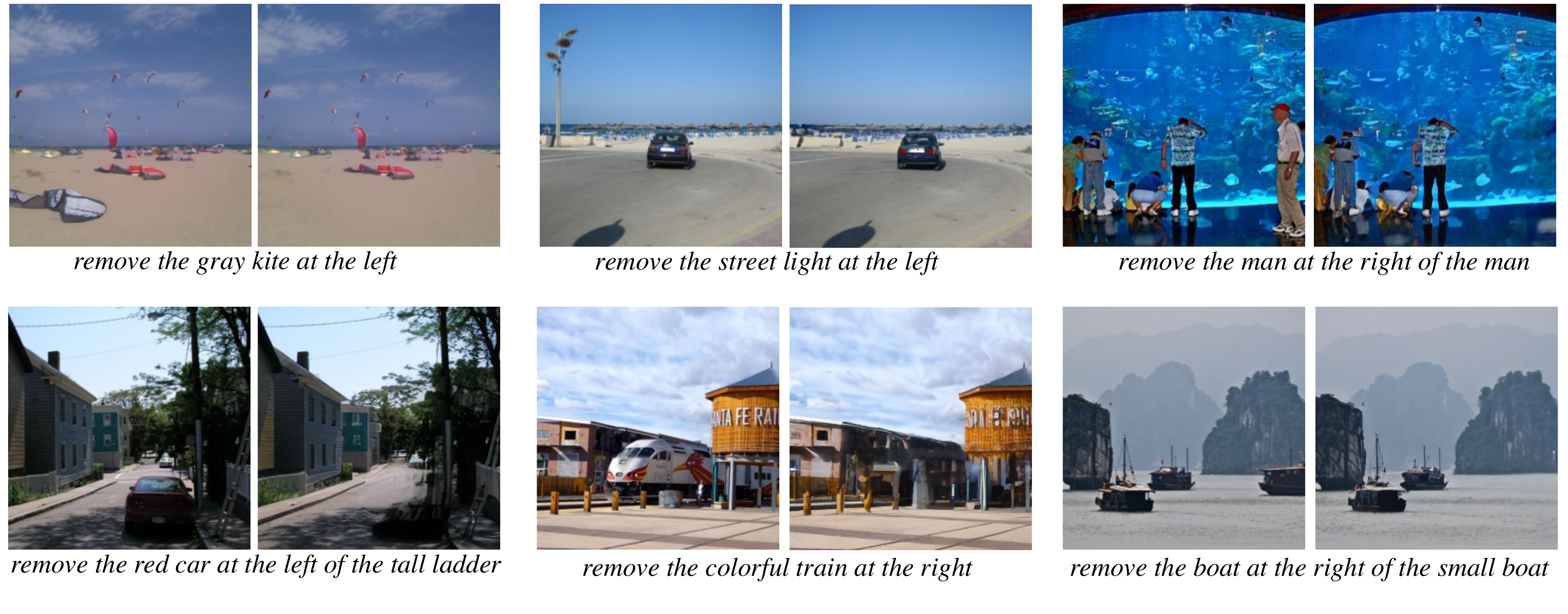}
    \caption{\textbf{Instructional Image Inpainting.} We propose a new image inpainting method, \textbf{Inst-Inpaint}, which takes an image and a textual instruction as input and is able to automatically remove objects mentioned in the text, without any need for a user-supplied binary mask to locate the region of interest, as in traditional inpainting approaches. We show sample results of our method trained and tested on a new real image inpainting benchmark dataset, \textbf{GQA-Instruct}, which we created for the proposed instructional image inpainting task.}
    \label{fig:teaser}
\end{figure*}

In this work, we introduce a novel task that we refer to as \emph{instructional image inpainting}, in which the unwanted objects are specified only via textual instructions, without any need for binary masks (see Fig.~\ref{fig:teaser}). To tackle this problem, by building upon the GQA dataset~\cite{hudson2019gqa}, we first use a data generation pipeline to create a new benchmark dataset, which we call GQA-Inpaint. Second, we design a single-stage deep inpainting network, \name, that can remove objects from images using instructions given as text prompts. Compared to existing inpainting methods, our model does not require masks, and neither do we predict masks explicitly. In summary, our contributions are as follows:
%
%
\begin{itemize}[noitemsep,leftmargin=*]
    \item We propose an end-to-end image inpainting framework, \name, that can perform the removal of an object in a given image based solely on textual instructions. Some example outputs of our method are shown in Fig.~\ref{fig:teaser}.
    \item We generate a real image dataset, GQA-Inpaint, to train and evaluate models for the proposed instructional image inpainting task. Here, we use the GQA dataset~\cite{hudson2019gqa} originally proposed for visual reasoning, and exploit the provided scene graphs to generate paired training data by utilizing state-of-the-art instance segmentation and inpainting methods.
    \item We conduct extensive experiments to show the effectiveness of our framework. To compare the results, we evaluate various baselines using several metrics, which include a novel CLIP~\cite{radford2021clip}-based inpainting score. We achieve significant improvements over the state-of-the-art for text-based image inpainting methods. 
\end{itemize}

\section{Related Work}

\vspace{0.1cm}\noindent\textbf{Diffusion-based Generative Models.} Diffusion models empowered with large-scale text-image paired datasets achieve impressive results on image synthesis \cite{nichol2021glide, saharia2022photorealistic, ramesh2022hierarchical, avrahami2022spatext, balaji2022ediffi}. The image synthesis capabilities of diffusion models are extended to video generation \cite{wu2022tune, singer2022make} and image editing \cite{meng2021sdedit, hertz2022prompt, avrahami2022spatext, song2022objectstitch, brooks2022instructpix2pix, singh2022high} tasks.
Our work tackles the inpainting task given text-based instructions and is more related to the editing tasks.
SDEdit is one of the first works that edits real images with  diffusion models \cite{meng2021sdedit}. It applies noise to input images modified with brush strokes and denoise them. Content preservation is not strictly achieved given the noising-denoising steps.
Prompt-to-prompt editing \cite{hertz2022prompt} shows via operations applied on the attention mechanisms, generated images can be edited with text prompts.
InstructPix2Pix \cite{brooks2022instructpix2pix} generates a dataset with Prompt-to-prompt and fine-tunes a diffusion model to edit images with paired data.


\vspace{0.1cm}\noindent\textbf{Image Inpainting.} The common set-up for the image inpainting task includes a binary mask that defines the erased pixels \cite{yu2019free, liu2022partial}. The original image pixels are removed and new ones are generated based on those masks. 
This domain was previously dominated by Generative Adversarial Networks (GANs) \cite{pathak2016context, liu2018image, yu2019free, li2020recurrent, liu2022partial, yildirim2023diverse}.
 GAN-based models mostly output deterministic results as these models are trained with reconstruction losses as well for stability \cite{yu2019free, liu2022partial, yu2022high}.
 There are also recent methods that achieve diversity with GANs  \cite{zhao2021large, li2022mat}.
 However, they are trained on single object datasets such as Faces.
 They are not extended to inpaint diverse scenes because GANs are notoriously difficult to train.
 
 Recently, diffusion-based models also entered this domain with promising results \cite{lugmayr2022repaint, rombach2022high}.
 It is shown that  a pretrained unconditional diffusion model can inpaint images by Repaint model \cite{lugmayr2022repaint}.
 Repaint modifies the denoising process to condition generation on the unerased image content.
Concurrent to our work is X-Decoder \cite{zou2022generalized} which can process an input image and text together for referring segmentation.
X-Decoder can segment images based on prompts and when combined with diffusion models, can erase the segmented objects.

\section{Dataset Generation}
\label{sec:dataset_method}

Our goal is to compose a real-image dataset to train and evaluate models for instruction-based image inpainting. There exist some datasets that can be adapted for this task like the CoDraw \cite{CoDraw} and the CLEVR \cite{johnson2017clevr} datasets. However, the images in these datasets all depict synthetic scenes composed of either geometric objects or clip arts. Although these relatively simple images can be regarded as great tools for building models, the models trained on them are not directly applicable to real image editing.

\begin{figure*}[!t]
    \centering
    \includegraphics[width=0.90\linewidth]{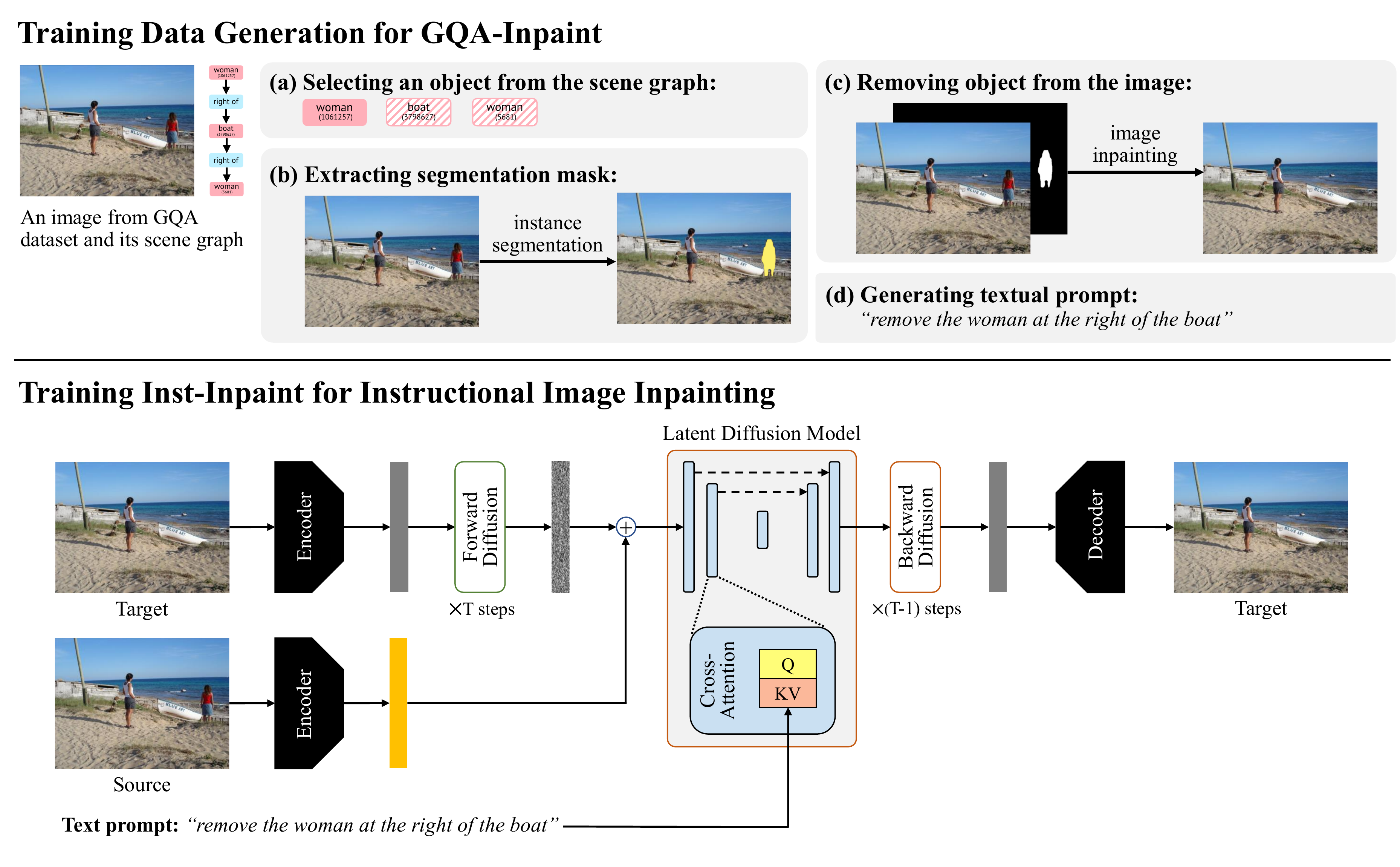}
    \caption{\textbf{The proposed GQA-Inpaint dataset and our Inst-Inpaint method.} Our work involves initially generating a dataset for the proposed \emph{instructional image inpainting} task. To create input/output pairs, we utilize the images and their scene graphs that exist in the GQA dataset~\cite{hudson2019gqa}. (a) We first select an object of interest. (b) We perform instance segmentation to locate the object in the image. (c) We apply a state-of-the-art image inpainting method to erase the object. (d) Finally, we create a template-based textual prompt to describe the removal operation. As a result, our \textbf{GQA-Inpaint} dataset includes a total of \textbf{147165} unique images and \textbf{41407} different instructions. Trained on this dataset, our \textbf{Inst-Inpaint} model is a text-based image inpainting method based on a conditioned Latent Diffusion Model~\cite{rombach2022high} which does not require any user-specified binary mask and performs object removal in a single step without predicting a mask, as in similar works.}
    \label{fig:dataset_gen}
\end{figure*}

To alleviate this drawback of the existing datasets, in this work, we propose to build a new image inpainting dataset that involves complex real images without using any human annotation. We refer to this dataset as \textit{\dataset} as we opt for building it on top of the GQA Dataset \cite{hudson2019gqa}. In particular, the GQA dataset consists of 85K real-world images with their corresponding scene graphs. Scene graphs provide simplified representations of images by representing them in terms of objects, attributes, and relations. Each node in a scene graph denotes an object, with its position and size specified by a bounding box. Moreover, each object is associated with a number of attributes encoding its color, shape, or material. The edges connecting the nodes and thus objects generally represent the spatial relations between these objects.


Our data generation pipeline for the proposed dataset shown in Fig. \ref{fig:dataset_gen} heavily depends on the scene graph representation. Specifically, the objects for the inpainting task are selected from the nodes of the scene graphs. Although the bounding boxes of the objects are already available in the GQA dataset, we observe that directly using them as masks for performing inpainting leads to unsatisfactory results. Hence, to obtain higher-quality target images, we extract segmentation masks of each object and use them for removing objects from images. Additionally, we generate the textual prompts used as the instructions from the provided scene graphs. In the following, we describe these steps.

\begin{figure}[!h]
    \centering
    \includegraphics[width=0.8\linewidth]{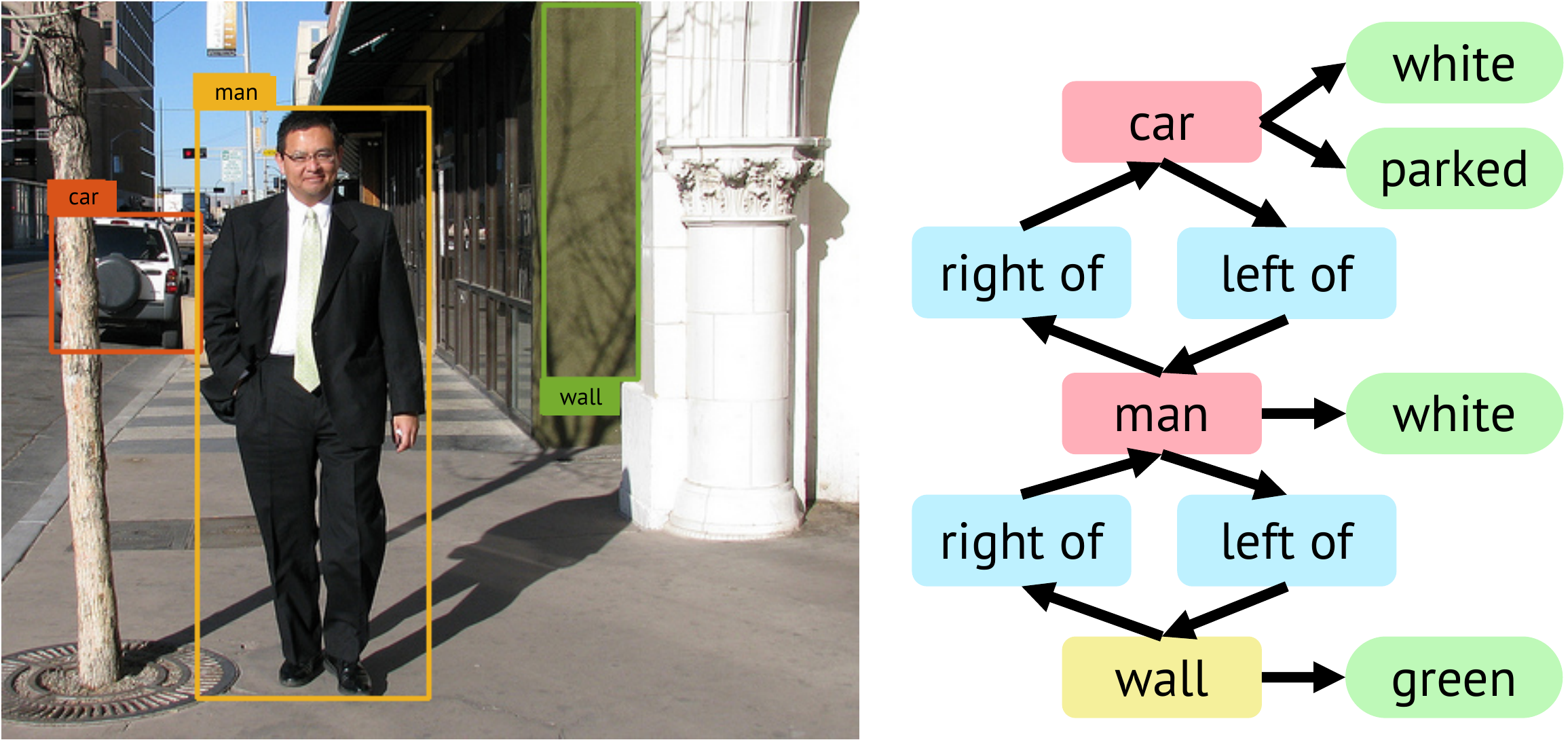}
    \caption{A sample image from the GQA dataset~\cite{hudson2019gqa} and the corresponding scene graph.}
    \label{fig:data-example}
\end{figure}

\vspace{0.1cm}\noindent\textbf{Selecting Objects from the Scene Graphs.}
In generating text prompts, we select among the objects included in the scene graphs in the GQA \cite{hudson2019gqa} dataset mainly based on the following two questions: \textit{``Is it reasonable to remove this object from the scene?''} and \textit{``Would the removal operation result in an unambiguous scene?''}. Following these questions, we can divide objects in the scene graphs into two broad categories: (i) objects that can be removed and can be uniquely referred to in relation to other objects such as man, boat, and kite, (ii) objects that can appear in referring expressions, but removing them does not make much sense e.g. wall, sky. 
We manually annotate scene graphs with an attribute to define these categories with a "bidirectional" flag.
From a scene graph point of view, it can be said that the objects in the first group have bidirectional relationships, while those in the second group have unidirectional relationships. For example, consider the image and its scene graph shown in Fig.~\ref{fig:data-example}. While the \textit{wall} may appear in the expression when referring to the \textit{man}, e.g. `\textit{the man to the left of the wall}', it cannot be removed from the scene.
On the other hand, man and car can be removed from the scene because their bidirectional flag is set to True.

Some of the scene graphs contain nodes that correspond not to a single object but to multiple objects from the same class, e.g. ``\textit{apples}", ``\textit{bikes}", ``\textit{windows}", etc. We manually detect them and opt for using these nodes not as target objects but as objects that can appear in a referring expression, identifying the object of interest.

We also filter out certain instances of the objects based on their sizes and their part-whole relations with other objects. Our criteria are as follows:
\begin{itemize}
    \item \textbf{Object Size}. We do not consider the objects that have an area greater than half of the whole image area as the objects to be removed. We notice that removing those large objects which occupy too much space from an image results in improbable inpainting results. 
    Additionally, we also discard the objects that are too small, which correspond to the objects covering an area smaller than 0.0025$\%$ of the whole image. We observe that inpainting such small objects results in scenes that are very much similar to the original scene. We decide these threshold values empirically based on our visual inspections. 
    \item \textbf{Parts vs. Objects}. We exclude the object classes that can be regarded as parts of some other objects from the list of objects that can be removed. In particular, these objects can be categorized into two as follows:
    \begin{itemize}
        \item \textit{Implicit parts of an object.} For example, \textit{leg}, \textit{arm}, and \textit{eye} for a human body, \textit{wheel} for a car, and \textit{tail} for a cat belong to this category. Inpainting these implicit parts from images results in non-realistic and visually poor scenes. 
        \item \textit{Items worn by an object.} For example, \textit{jacket}, \textit{pants}, \textit{shirt}, \textit{jeans}, \textit{shoes} worn by people belong to this category. Like the implicit parts, inpainting these wearable items gives implausible results, while also raising some ethical considerations. 
    \end{itemize}
\end{itemize}

\begin{figure}[!t]
    \includegraphics[width=1\linewidth]{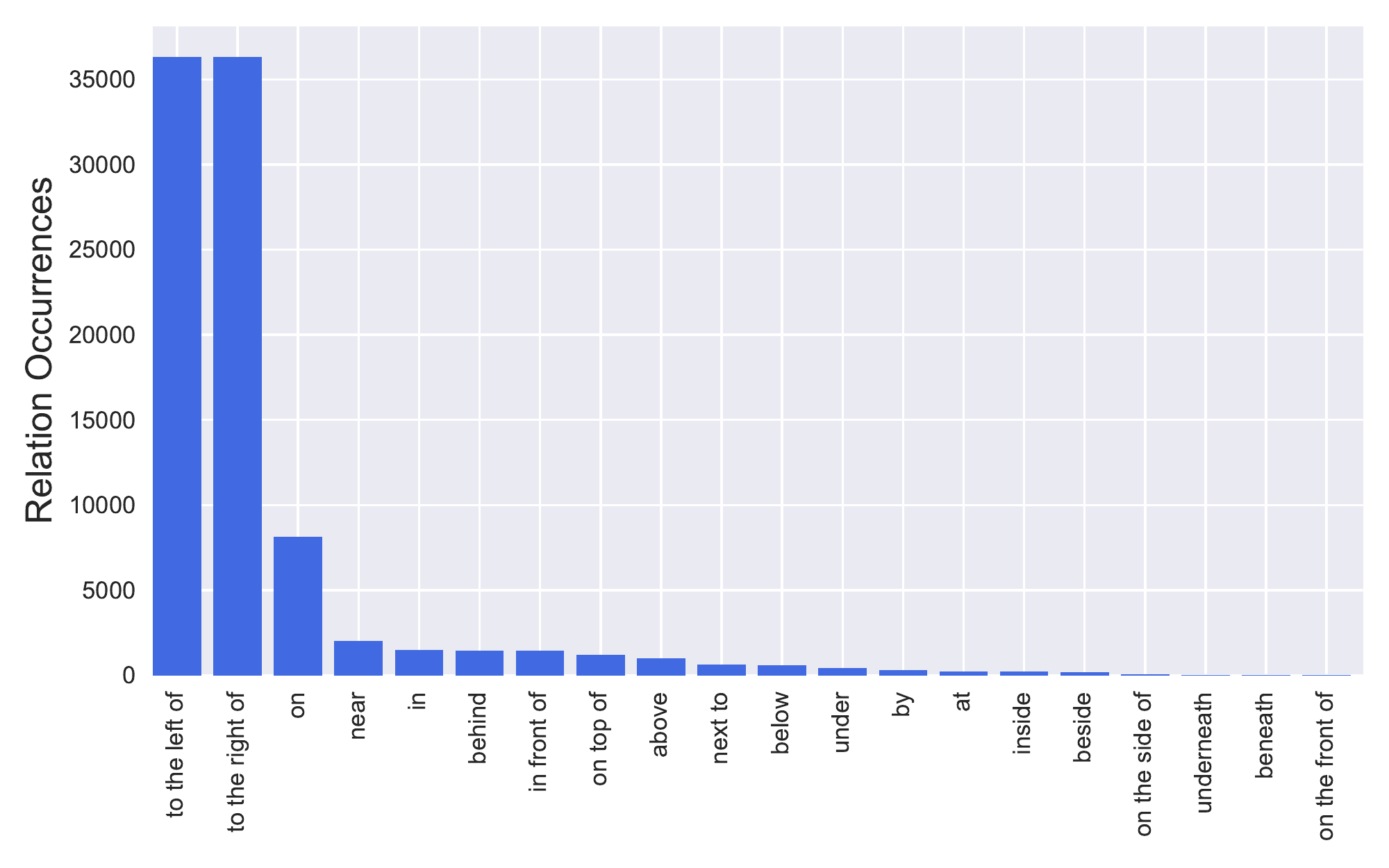}
    \caption{Distribution of the relation types exist in the proposed GQA-Inpaint dataset (sorted by their number of occurrences).}
    \label{fig:relation_occ}
\end{figure}

\vspace{0.1cm}\noindent\textbf{Selecting Relations from the Scene Graphs.}
Instead of using all the relations that exist in the scene graphs, we remove the rarely seen relations that have a frequency rate lower than 0.0001. After this filtering operation, we
end up with the relations provided in Fig. \ref{fig:relation_occ}.
Additionally, some objects in the scene graph do not have any defined relation with any other object. 
We follow a simple rule for generating relations for this kind of objects by using their spatial location. Specifically, we split an image into three equal sections along the x-axis, and name these sections as ``\textit{left}", ``\textit{center}", and ``\textit{right}", following the natural naming. We generate the relation descriptions in the form of ``\textit{at the \emph{[location]}}'' with the \textit{\emph{[location]}} referring to one of the aforementioned sections that largely covers the object of interest.

\vspace{0.1cm}\noindent\textbf{Extracting Segmentation Masks.} We obtain segmentation masks of objects with Detectron2 \cite{wu2019detectron2} and Detic \cite{zhou2022detecting} frameworks. To cover a wide range of object categories, we segment objects with multiple instance segmentation models  that are trained on COCO \cite{lin2014microsoft} and LVIS dataset \cite{gupta2019lvis}.

\begin{itemize}
    \item \textbf{Detectron2} \cite{wu2019detectron2}. We extract the segmentation masks using the models available in Detectron2 Model Zoo, namely Mask R-CNN \cite{he2017mask}, Cascade R-CNN \cite{cai2019cascade}, and Panoptic FPN \cite{kirillov2019panoptic}. The class labels that these models trained on are, in general, different from the class labels from the GQA dataset. To find the proper class label of an object, we compare the Intersection over Union (IoU) scores of the prediction bounding box with all of the bounding boxes objects from the scene graph, and label the object with the class label with the highest IoU score. 
    \item \textbf{Detic} \cite{zhou2022detecting}. We use SwinB-21K \cite{liu2021Swin} model trained on LVIS \cite{gupta2019lvis}, COCO \cite{lin2014microsoft} and ImageNet-21K\cite{deng2009imagenet} datasets to extract the segmentation masks. In addition to an input image, the Detic framework also expects a set of possible class labels to predict the segmentation masks. For that, we consider the list of object classes available in the COCO dataset along with our own custom-made vocabularies extracted by making use of the attributes of the objects along with the object names (e.g., ``\textit{red chair}"), and the object names alone (e.g., ``\textit{chair}"). To retrieve segmentation masks whose vocabulary differs from ours, we follow the same methodology explained for Detectron2 reviewed above.
\end{itemize}

\begin{figure*}[!t]
    \centering
    \includegraphics[width=\linewidth]{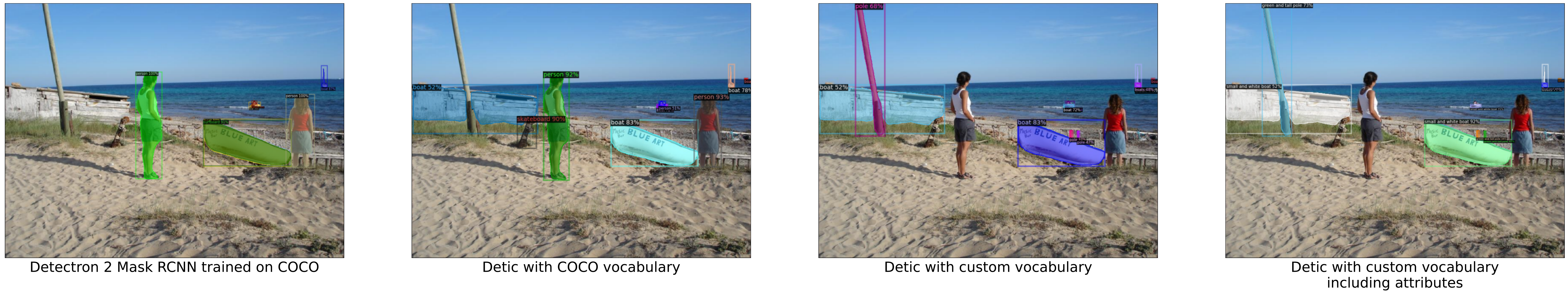}
    \caption{Comparison of Mask Extraction methods. The vocabulary of the COCO dataset is different from the vocabulary of the GQA dataset. To extract the correct segmentation masks, we make use of a combination of pretrained models available in Detectron2's Model Zoo and use the vocabulary generated based on the scenes to predict the corresponding class labels using the Detic2 framework.
    Different setups provide us with more options to pick the most accurate instance segmentation mask.}
    \label{fig:mask-extract}
\end{figure*}

Fig. \ref{fig:mask-extract} shows the instance segmentation masks predicted by the aforementioned methods and setups. By combining the results of different models, we obtain more accurate segmentation masks for a notable number of object classes. Among these predictions by the multiple methods and setups, we pick the instance segmentation masks having higher IoU between ground-truth and detection bounding boxes as the final object mask.

\begin{figure*}[!t]
    \centering
    \includegraphics[width=0.99\linewidth]{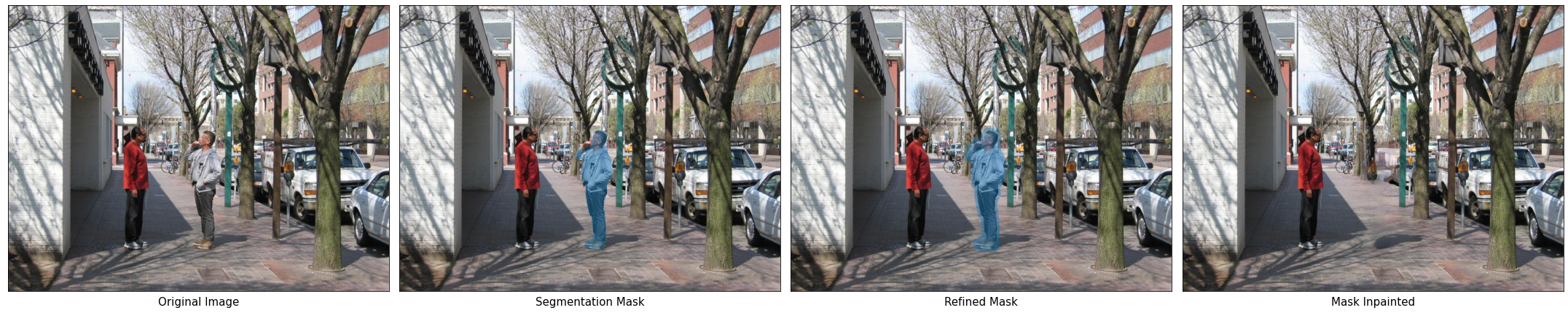}
    \caption{An illustration of the intermediate steps of the object removal operation that we considered while constructing GQA-Inpaint. While the first column depicts the original image, the second row shows the segmentation mask detected by a pretrained instance segmentation model. The third column demonstrates the object mask used for inpainting, which is first refined by the CascadePSP~\cite{cheng2020cascadepsp} and then expanded by a dilation operation. The fourth column presents the final inpainting result obtained by~CR-Fill \cite{zeng2021cr}.}
    \label{fig:sup_dataset_gen}
\end{figure*}

\vspace{0.1cm}\noindent\textbf{Removing Objects From Images.} Our goal in this step is to remove the objects from the image. For that purpose, we use a state-of-the-art image inpainting method, CRFill \cite{zeng2021cr} due to its computational efficiency and high-quality results. Before performing inpainting, we first apply CascadePSP \cite{cheng2020cascadepsp} to refine the predicted masks and then use morphological dilation with a structuring element of 11$\times$11 pixels to expand the segmentation masks at the edges, to make the masks better capture the objects. These steps are demonstrated in Fig. \ref{fig:sup_dataset_gen}. 

\begin{figure*}[!t]
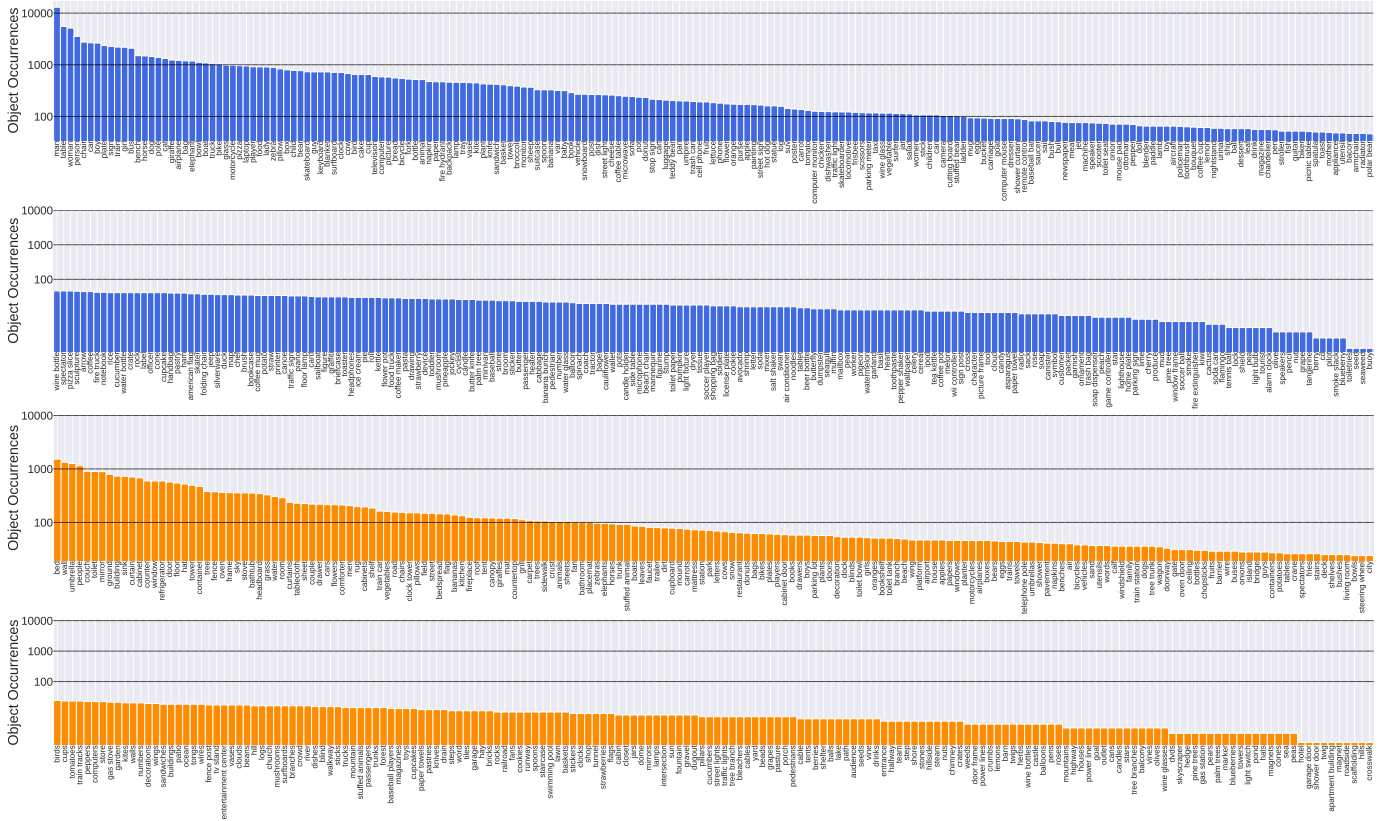

    \centering
    \includegraphics[width=\linewidth]{figures/dataset/object_occurrences/new/occ1_1.pdf}
    \includegraphics[width=\linewidth]{figures/dataset/object_occurrences/new/occ1_2.pdf}
    \includegraphics[width=\linewidth]{figures/dataset/object_occurrences/new/occ1_3.pdf}
    \includegraphics[width=\linewidth]{figures/dataset/object_occurrences/new/occ1_4.pdf}
    \caption{Distribution of the object classes exist in the proposed GQA-Inpaint dataset,  sorted by their number of occurrences and categorized based on their roles. Objects that can be removed and can be referred to in relation to other objects are shown in blue color, while the objects that can appear in referring expressions are given in orange color. Please zoom in for text. }
    \label{fig:data-occur}
\end{figure*}

\vspace{0.1cm}\noindent\textbf{Generating Textual Prompts.} The GQA dataset contains relational representations between objects in scene graphs as shown in Fig. \ref{fig:dataset_gen}. 
If the selected object is a single instance of its category, we do not use any relations and simply create the prompts as ``\textit{remove the [object]}".
If multiple instances of the same object category are present in the image, we construct the prompt based on the relations from the scene graph to specify a specific instance.
For the object to be removed, we use all of its relations if they are available. Otherwise, we use the image relative position. Multiple relations are combined into a single relation by concatenating all of them with  ``\textit{and}''. For example, relations ``\textit{to the left of the man}'', and ``\textit{on the table}'' combined into a single relation ``\textit{to the left of the man and on the table}''.
In addition to the relations, we also make use of the object attributes both for the object to be removed and for objects that participate in the relations. Unlike relations, we do not combine all the available attributes. Instead, we randomly choose an attribute as a data augmentation method.
We provide the the object statistics in our dataset in Fig. \ref{fig:data-occur}.

\section{Method}
To solve the proposed instructional image inpainting task, we develop a novel conditional diffusion model named \textit{\name } which takes an image and a textual instruction as inputs. Training this model requires a paired dataset that includes source and target images with the corresponding text prompts. 
We build our model based on the latent diffusion model \cite{rombach2022high} due to its computational efficiency. Specifically, the latent diffusion model includes an encoder $E$, and decoder $D$ to project images to a lower latent dimension and to reconstruct the images back, respectively. It is a variational autoencoder and is trained separately as the first step.

The diffusion process \cite{ho2020denoising} takes place on the latent codes, $z_0 = E(x)$, with $x$ denoting the input image.
Noise is gradually added to $z_0$ at each time step $t$ such that after $T$ number of steps, $z_T \sim N(0, I)$.
Diffusion models learn denoising auto-encoders which are trained to reverse the process and predict a denoised variant of their input, $z_t$ where $z_t$ is a noisy version of $z_0$.
This objective that operated on the latent codes is as follows:
\begin{equation}
    L_{LDM} := \mathbb{E}_{E(x), \epsilon \sim N(0, 1), t}[ || \epsilon - \epsilon_\theta (z_t, t) || ]
\end{equation}
\noindent with $t$ sampled from $1, .... , T$ and $\epsilon_\theta (., t)$ is a neural network that predicts the added noise conditioning on the time-step $t$.
Similar to previous works \cite{rombach2022high}, we use a U-net architecture.  

As stated, our model for instructional image inpainting can be interpreted as a conditional diffusion model. That is, we train our model with two additional conditioning;  source image and text prompt. During training, the target image goes through the encoder to obtain $z_0$ as shown in Fig. \ref{fig:dataset_gen}. The source image, $s$, is also encoded with the same pre-trained encoder from the first stage. The encoded features, $E(s)$ are concatenated with $z_t$ at each time step. For the text conditioning, we use the  cross-attention mechanism \cite{vaswani2017attention}.
The objective of the instruction-based object removal is given as follows:
\begin{equation}
    L_{LDM} := \mathbb{E}_{E(x), \epsilon \sim N(0, 1), t}[ || \epsilon - \epsilon_\theta (z_t, t, E(s), T(i)) || ]
\end{equation}
\noindent Here, $i$ refers to the text instruction and $T$, the transformer. The encoder, $E$, is fixed during training whereas $\epsilon_\theta$ and $T$ are optimized together. 

\textbf{Implementation and Training Details.}
We train two Inst-Inpaint models; one on the GQA-Inpaint and one on the CLEVR datasets. 
The reason for us to train a model on the CLEVR dataset is to compare the Inst-Inpaint model with GAN-based works.
Because we were not able to train the previous instruction-based image-erasing works that are built on GANs successfully on the real GQA-Inpaint dataset. Therefore, we compare with them on the CLEVR dataset. More discussion on this comparison is given in Sec. \ref{sec:baseline}.
In Table~\ref{tab:hyperparameters}, we provide an overview of the hyperparameters.  While the GQA-Inpaint model uses a pretrained VQGAN model\footnote{https://github.com/CompVis/taming-transformers} as the first stage encoder, we train an autoencoder with KL regularization for the CLEVR model. 
For both stages of the training, center cropping followed by resizing is applied to both datasets.



Our model has two types of conditioning in the second stage, which are source image and text. For source image conditioning, we concatenate the latent representation of the source image with the noisy image at the current timestep. The latent representation is obtained by using the first stage encoder, which is frozen during the training phase. Therefore, the U-Net architectures have doubled latent channel sizes for the input data. For text conditioning, we apply cross-attention between the text embeddings and the specific U-Net activation layers. For this purpose, as in the text-to-image model of LDM, we train a Transformer model \cite{vaswani2017attention} from scratch together with the U-Net by feeding BERT \cite{devlin2018bert} tokens as input to extract in-domain semantic representations from the inpainting instructions.


\begin{table}[!t]
    \centering
    \resizebox{0.9\linewidth}{!}{
    \begin{tabular}{lcc}
        \toprule
        & CLEVR & GQA-Inpaint\\
        \midrule
        First Stage Encoder & & \\
        $\quad$Regularization Type & KL & VQ\\
        $\quad$$z$-shape & $16\times16\times4$ & $32\times32\times4$\\
        $\quad$$|\mathcal{Z}|$ & $-$ & $16384$\\
        $\quad$Epochs & $400$ & Pretrained\\
        $\quad$Self-attention Res. & $16$ & $32$ \\
        \midrule
        Conditional LDM & & \\
        $\quad$Diffusion Steps & $1000$ & $1000$ \\
        $\quad$Noise Schedule & linear & linear \\
        $\quad$Channels & $64$ & $128$ \\
        $\quad$Depth & $2$ & $2$ \\
        $\quad$Channel Multiplier & $1,2,4$ & $1,2,3,4$\\
        $\quad$Number of Heads & $4$ & $8$ \\
        $\quad$Batch Size & $16$ & $32$ \\
        $\quad$Epochs & $1000$ & $300$ \\
        $\quad$Learning Rate & $2.e\textrm{-}6$ & $2.e\textrm{-}6$ \\
        \midrule
        Text Cond. Cross Attn & & \\
        $\quad$Resolution & $32,64,128$ & $32,64,128$ \\
        $\quad$Transformer Depth & $1$ & $1$ \\
         Text Cond. Embedding & & \\
        $\quad$Dimension & $512$ & $512$ \\
        $\quad$Transformer Depth & $8$ & $16$ \\
        \bottomrule
    \end{tabular}}
    \caption{Hyperparameters for the proposed Inst-Inpaint model trained on the CLEVR and the GQA-Inpaint datasets.}
    \label{tab:hyperparameters}
\end{table}



\section{Experiments}

\subsection{Baselines}
\label{sec:baseline}

We compare Inst-Inpaint against diffusion and GAN-based models.
Our diffusion-based comparison models are Instruct X-Decoder \cite{zou2022generalized}, CLIPSeg \cite{luddecke2022image}, and InstPix2Pix \cite{brooks2022instructpix2pix}. 
Among these models, Instruct X-Decoder and ClipSeg first segment the referred objects. 
Later, it can be combined with image inpainting and editing models to achieve different applications.
We experiment with the released application of Instruct X-Decoder\footnote{https://huggingface.co/spaces/xdecoder/Instruct-X-Decoder} which combines X-Decoder for image understanding, GPT-3 for language understanding, and stable diffusion for image generation, and works in a zero-shot manner. It is important to state that all these components are trained on large-scale datasets which are much larger than our generated GQA-Inpaint dataset.
We additionally experiment with ClipSeg by replacing Instruct X-decoder's segmentation detection with the ClipSeg detections.
InstPix2Pix is similar to our method which is trained end-to-end for translation images based on instructions.
It is trained on a paired image dataset that is generated by the authors. 
The paired images based on the instructions are generated with prompt-to-prompt method \cite{hertz2022prompt}.

Currently, image generation and editing methods that are trained on large-scale datasets usually employ diffusion models.
We also try comparing our model with GAN-based models such TIM-GAN~\cite{zhang2021text} and cMANIGAN~\cite{fan2022target}.
TIM-GAN learns a mask for object removal by encoding the input image and text. 
Encoded features are multiplied with the mask in a pixel-wise manner and an image is reconstructed via a decoder. The network is trained with adversarial losses. Additionally, reconstruction loss is applied to the mask prediction.
cMANIGAN also predicts a mask and learns to edit images with adversarial losses and reconstruction loss on the mask. cMANIGAN proposes to use a cycle consistency loss by editing images twice; first with the given instruction and second with the reverse of it.
These models are tested on synthetic datasets. 
We train these models on our dataset but do not achieve any meaningful results. 
For the completeness of our study, we compare our method with these methods on a synthetic dataset; namely CLEVR dataset.

\subsection{Datasets}
\vspace{0.1cm}\noindent\textbf{GQA-Inpaint Dataset.}
The construction of this dataset is extensively explained in Section \ref{sec:dataset_method}.
We train the models on unique 44500 source images and unique 88369 target images.
Because from source images we pick different objects and generate instructions, training includes 173715 unique source-target-prompt  pairs.
We test the models on 4811 unique source images, 9485 unique target images, and 18883 unique source-target-prompt pairs.

\vspace{0.1cm}\noindent\textbf{CLEVR Dataset.} This dataset is generated by CLEVR toolkit \cite{vo2019composing, johnson2017clevr}. 
The dataset contains 3-D synthesized simple shapes with different colors and sizes. 
Each training pair includes a source image, a target image, and a text specifying the modification from the source image to the target image.
There are three types of text; add, remove, and change an object.
We train our model with the remove instructions. 
To increase the size of the dataset, we use the pairs with both remove and add instructions.
We flip the source-target pairs of add instructions and modify the text accordingly for the training dataset. 
This subset includes 5404 training and 2362 test images.

\begin{figure}
    \centering
    \includegraphics[width=\linewidth]{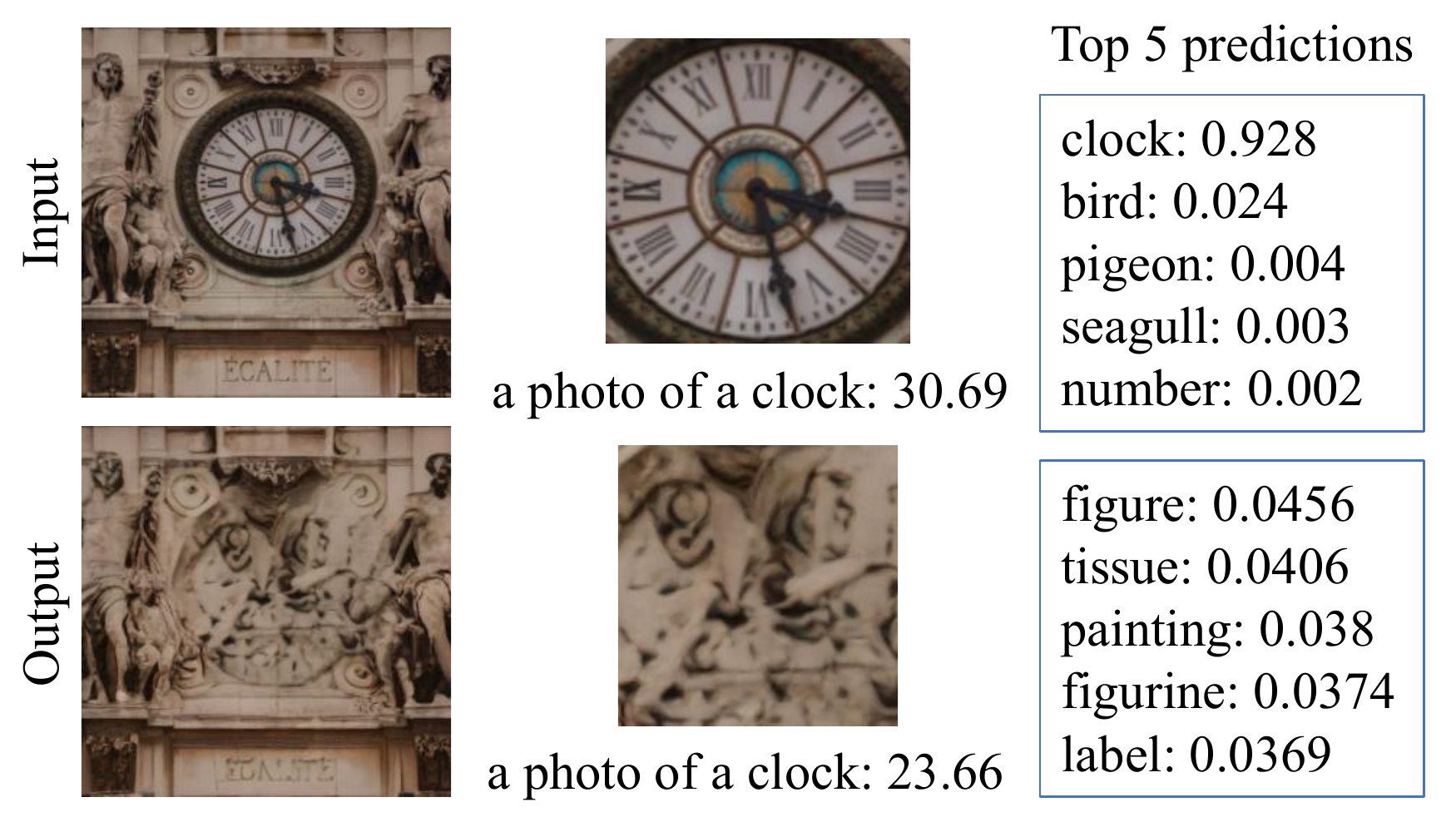}
    \caption{\textbf{CLIP-based image inpainting scores}. We use the image-text similarity scores and zero-shot classification scores in our evaluations. Please refer to Section \ref{sec:eval} for score descriptions.}
    \label{fig:clip_score}
\end{figure}

\newcommand{\interpfigk}[1]{\includegraphics[trim=0cm 0cm 0cm 0cm, clip, width=1.7cm]{#1}}
\newcommand{\interpfigtprompth}[1]{\includegraphics[trim=0 1cm 0 1cm, clip, width=2.0cm]{#1}}
\begin{figure*}[]
\centering
\addtolength{\tabcolsep}{-5pt}   
\begin{tabular}{cccccccccc}
Source & InstPix2Pix & X-Decoder & CLIPSeg & Ours & Source & InstPix2Pix & X-Decoder & CLIPSeg & Ours  \\
\interpfigk{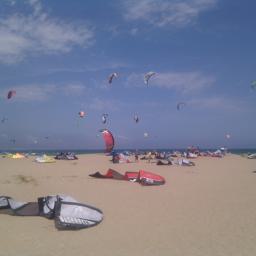} &
\interpfigk{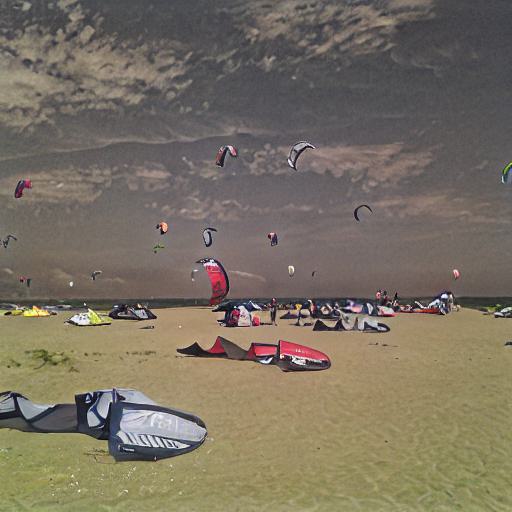} &
\interpfigk{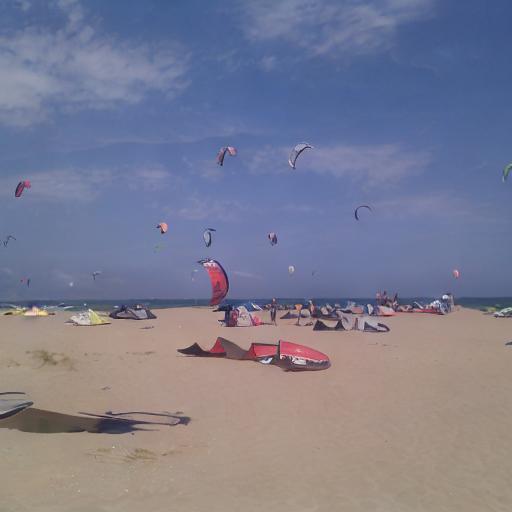} &
\interpfigk{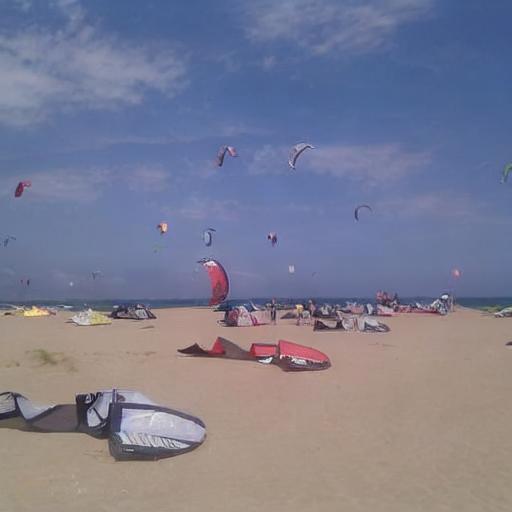} &
\interpfigk{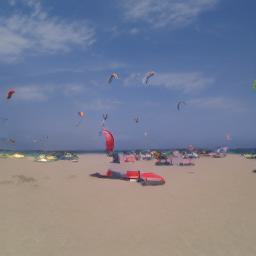} &

\interpfigk{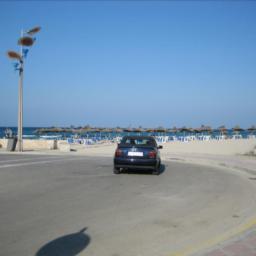} &
\interpfigk{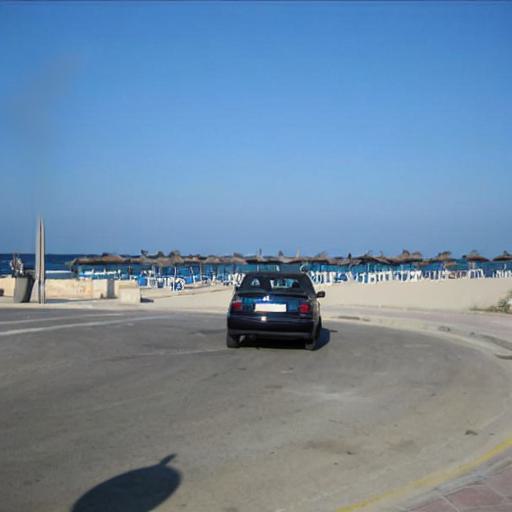} &
\interpfigk{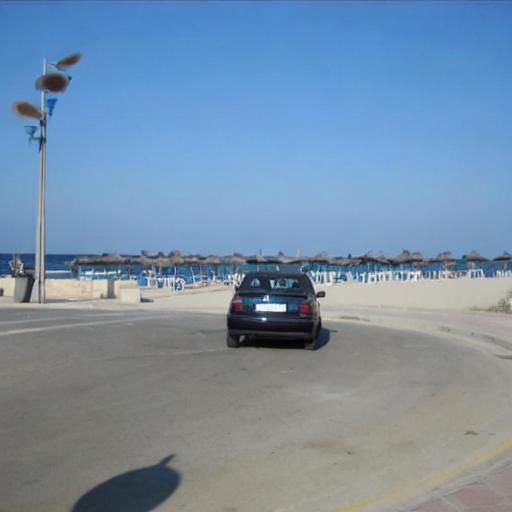} &
\interpfigk{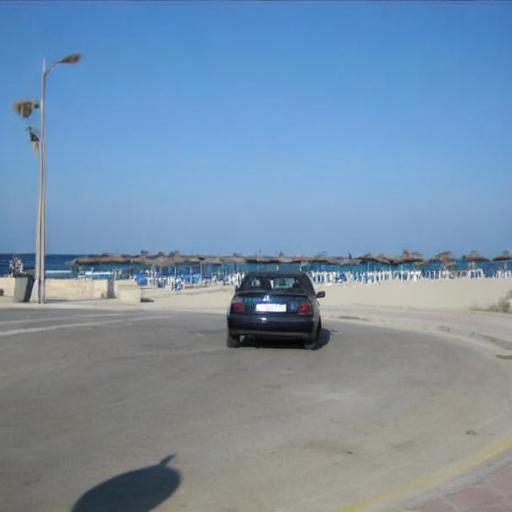} &
\interpfigk{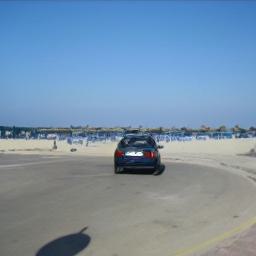}

\\
\ding{229} Remove & \multicolumn{4}{l}{{the gray kite}} & 
\ding{229} Remove & \multicolumn{4}{l}{{the street light at the left}}
\\

\interpfigk{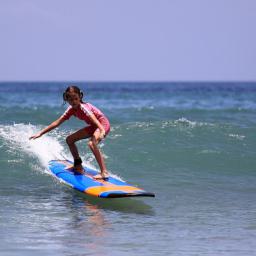} &
\interpfigk{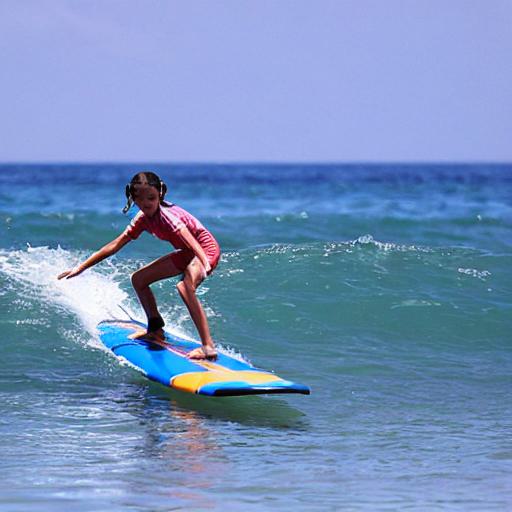} &
\interpfigk{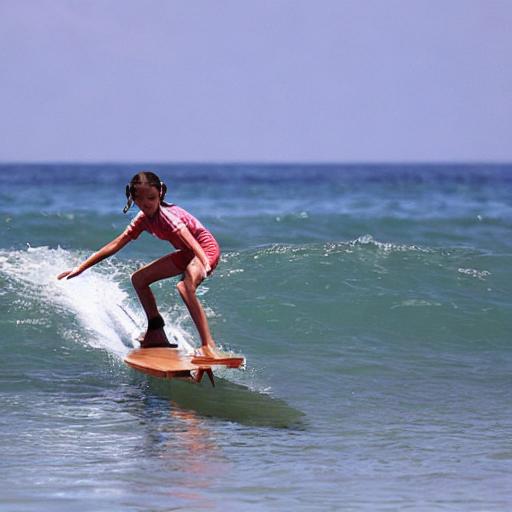} &
\interpfigk{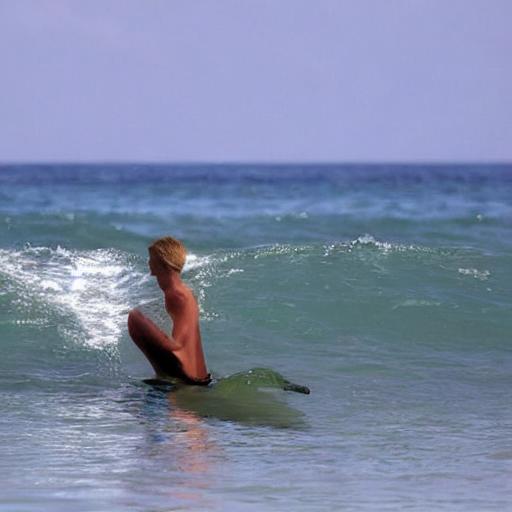} &
\interpfigk{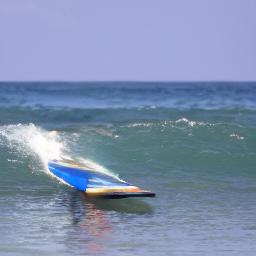} &

\interpfigk{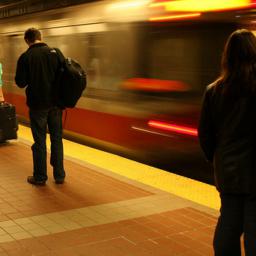} &
\interpfigk{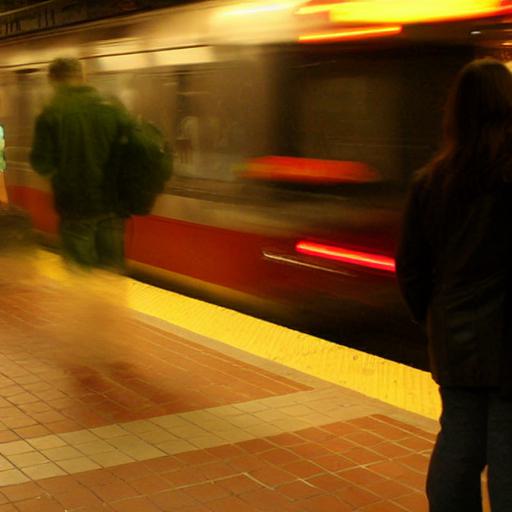} &
\interpfigk{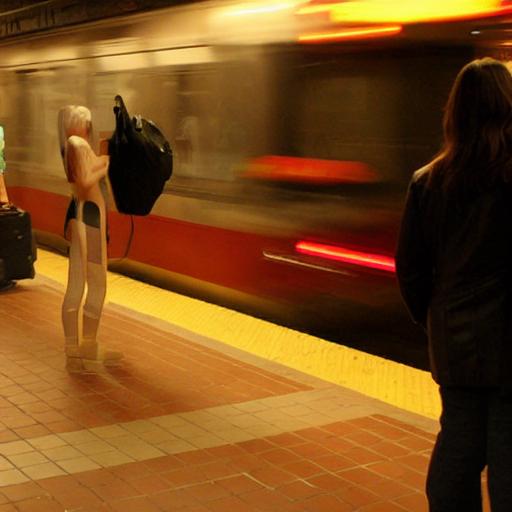} &
\interpfigk{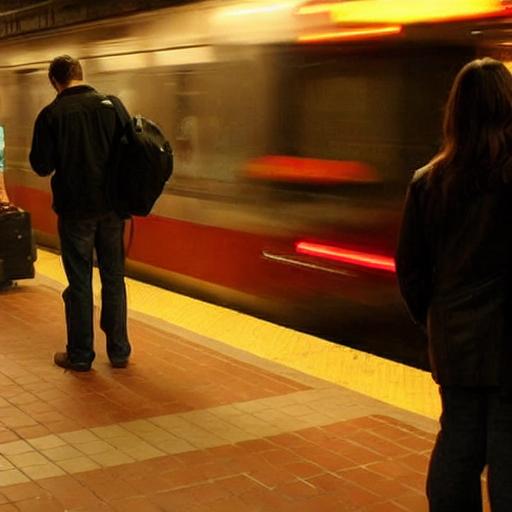} &
\interpfigk{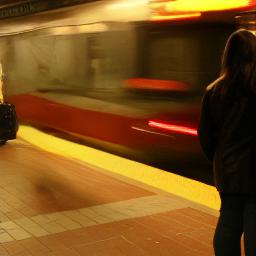}

\\
\ding{229} Remove & \multicolumn{4}{l}{{the surfing girl on the blue surfboard}} & 
\ding{229} Remove & \multicolumn{4}{l}{{the boy at the left}}
\\

\interpfigk{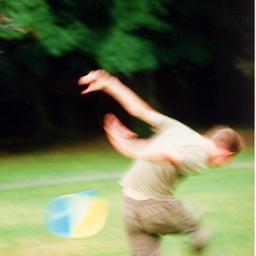} &
\interpfigk{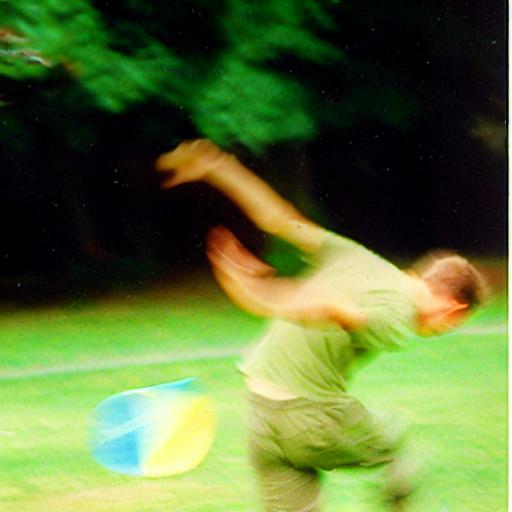} &
\interpfigk{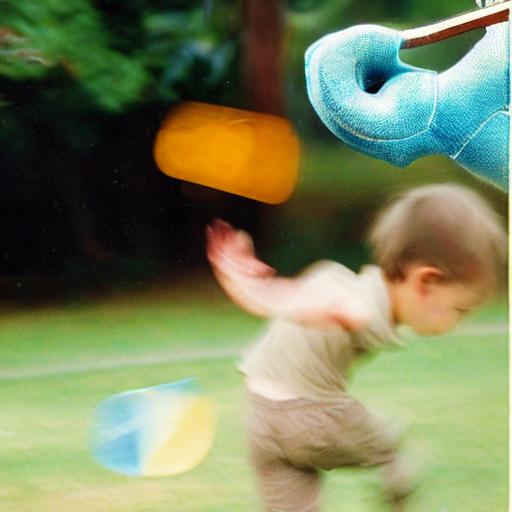} &
\interpfigk{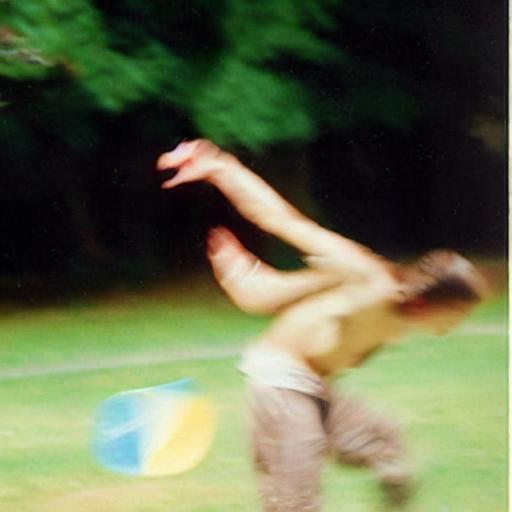} &
\interpfigk{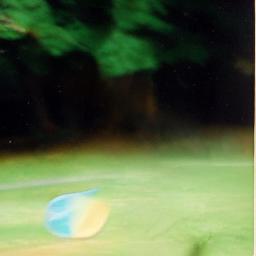} &

\interpfigk{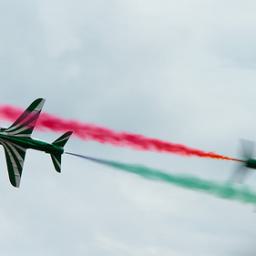} &
\interpfigk{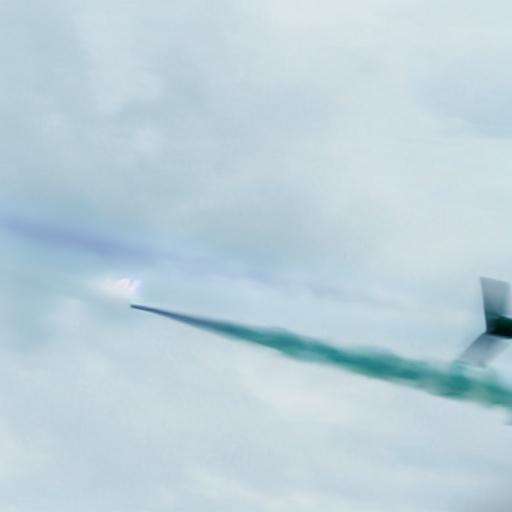} &
\interpfigk{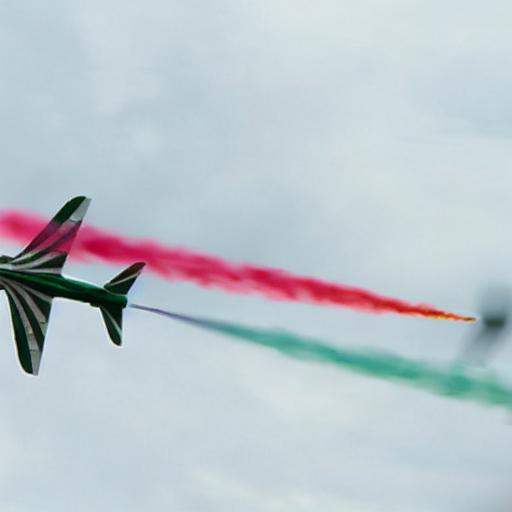} &
\interpfigk{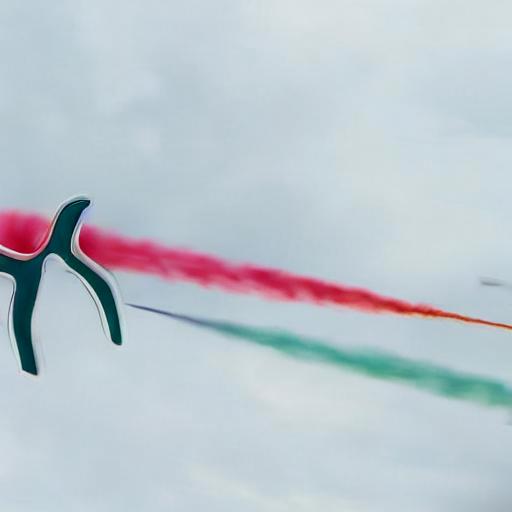} &
\interpfigk{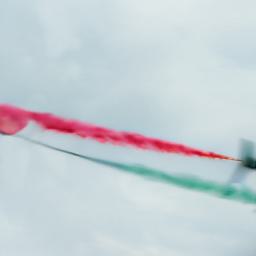}

\\
\ding{229} Remove & \multicolumn{4}{l}{{the boy}} & 
\ding{229} Remove & \multicolumn{4}{l}{{the airplane at the left}}
\\

\interpfigk{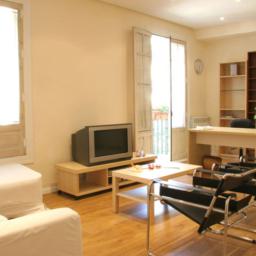} &
\interpfigk{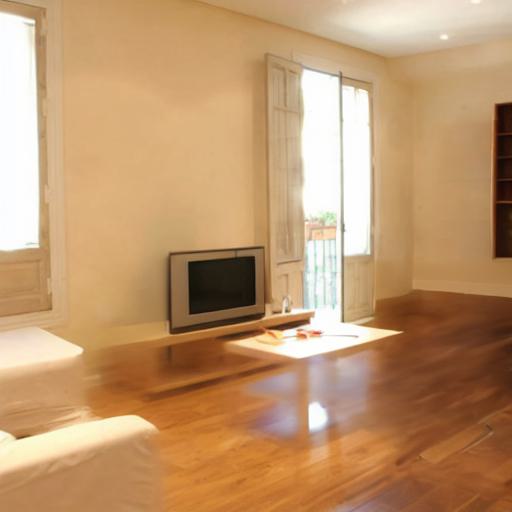} &
\interpfigk{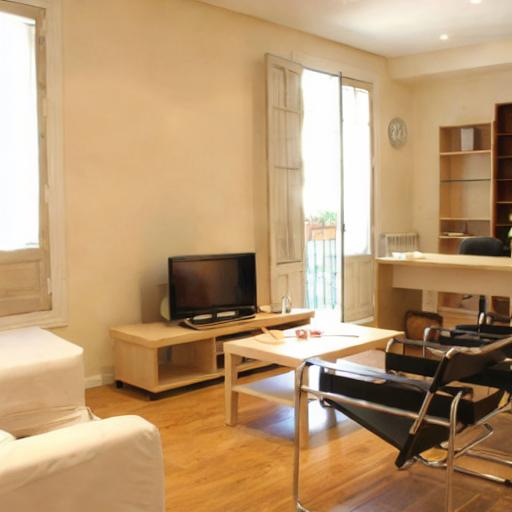} &
\interpfigk{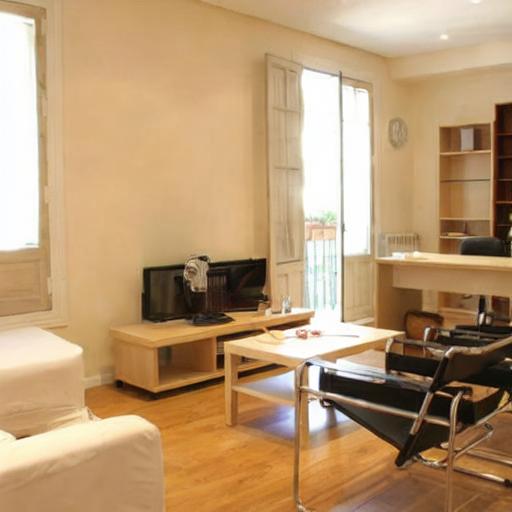} &
\interpfigk{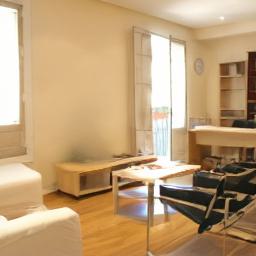} &

\interpfigk{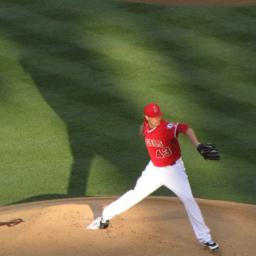} &
\interpfigk{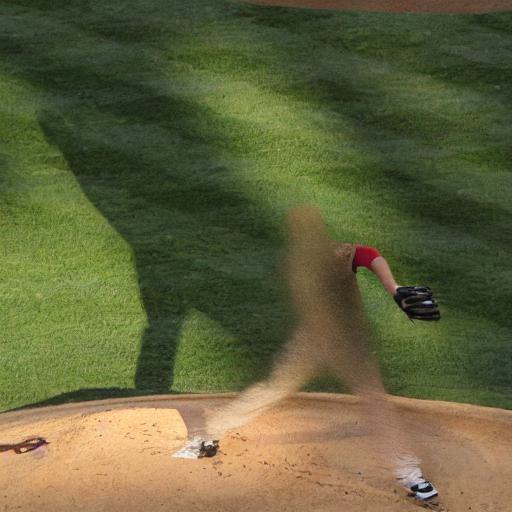} &
\interpfigk{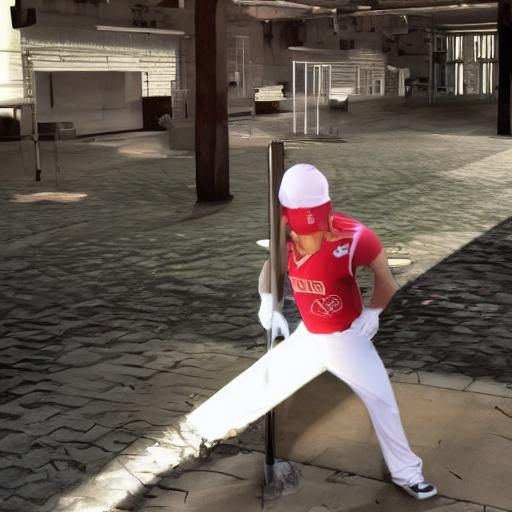} &
\interpfigk{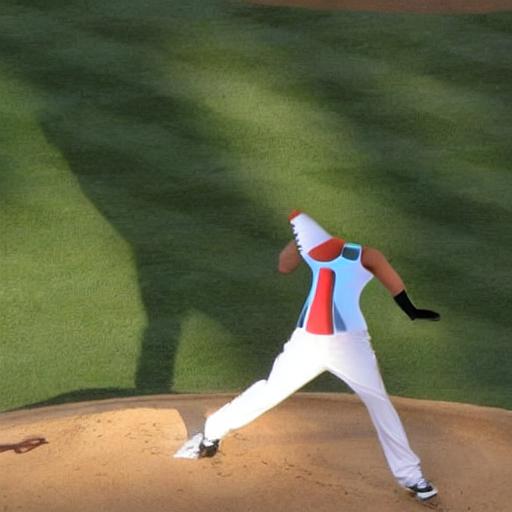} &
\interpfigk{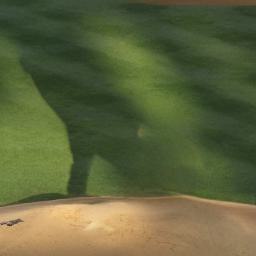}

\\
\ding{229} Remove & \multicolumn{4}{l}{{the television at the center}} & 
\ding{229} Remove & \multicolumn{4}{l}{{the man}}
\\

\interpfigk{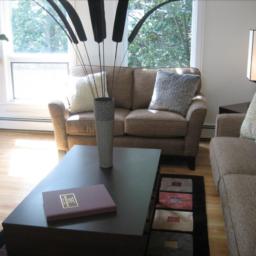} &
\interpfigk{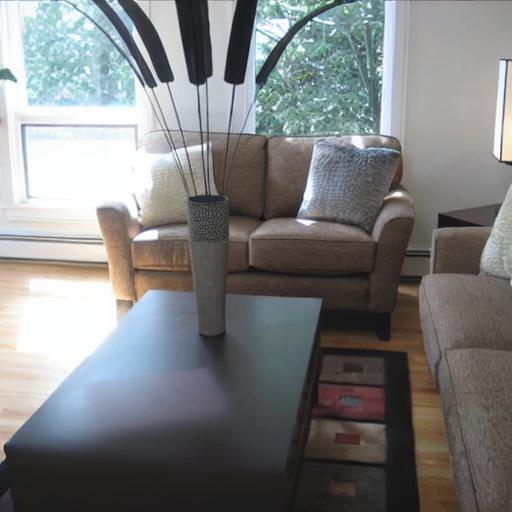} &
\interpfigk{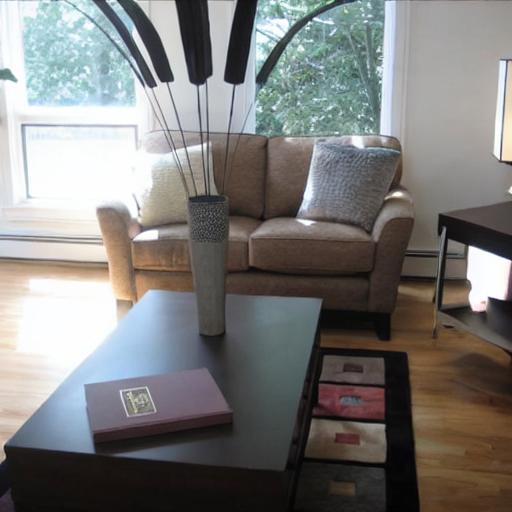} &
\interpfigk{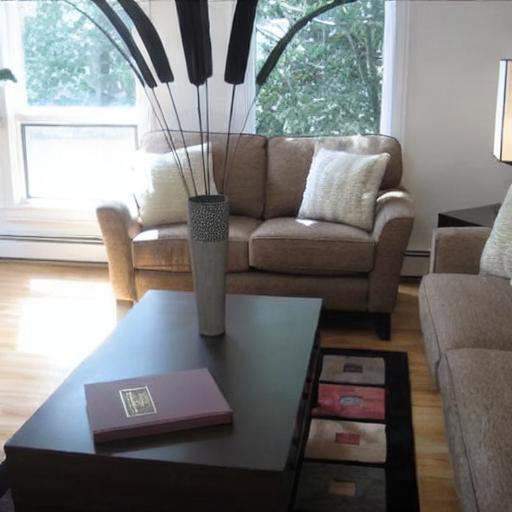} &
\interpfigk{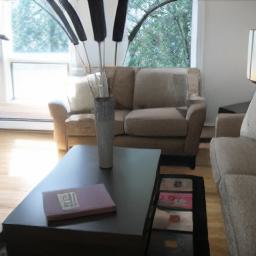} &

\interpfigk{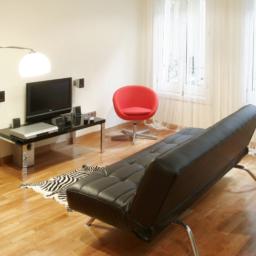} &
\interpfigk{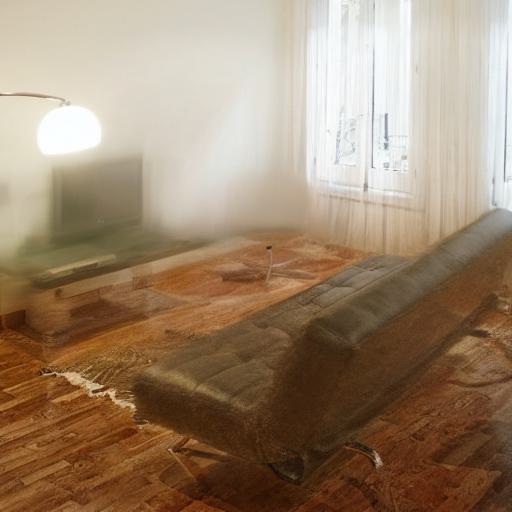} &
\interpfigk{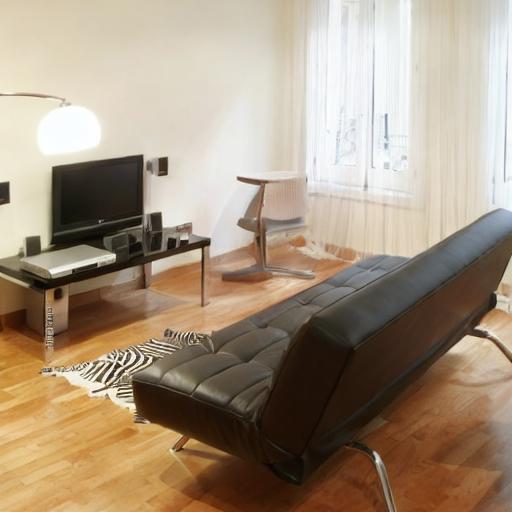} &
\interpfigk{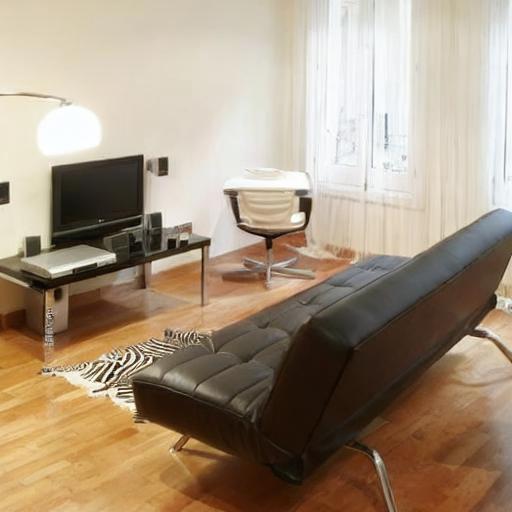} &
\interpfigk{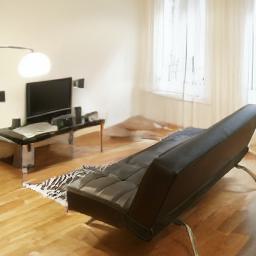}

\\
\ding{229} Remove & \multicolumn{4}{l}{{the pillow at the right}} & 
\ding{229} Remove & \multicolumn{4}{l}{{the red chair at the center}}
\\

\end{tabular}

\caption{Qualitative results of our approach and the competing methods on GQA-Inpaint dataset. Other methods sometimes struggle to detect the correct object and to erase the object completely. On the other hand, ours achieve significantly better results.}
\label{fig:results_real}
\end{figure*}



\subsection{Evaluation Metrics}
\label{sec:eval}

\noindent\textbf{FID.} We report the Frechet Inception Distance (FID) metric \cite{heusel2017gans}, which is commonly used for assessing the photorealism of the generated images by comparing the target image distribution and edited image distributions. For our instructional image inpainting task, removing the target object stated in the text prompt fully from the provided image is of importance. FID metric does not directly measure the accuracy of this removal operation, hence, to measure that, we additionally use CLIP-based measures and RelSim.

\vspace{0.1cm}\noindent\textbf{CLIP Distance.} The goal of CLIP distance is to evaluate how well the target object in the instruction is removed.
We use the bounding boxes linked to the nodes of the scene graphs for this metric. Specifically, our proposed \dataset~dataset inherently includes information regarding the bounding box of objects that are instructed to be removed and their semantic category labels. 
We extract the image regions as suggested by these bounding boxes from both the source and the inpainted images. 
We then estimate the CLIP similarities~\cite{radford2021clip}. between these extracted regions and the text prompts including semantic labels as shown in Fig. \ref{fig:clip_score}.
As in the example, we use the text prompt ``\emph{a photo of a clock}" to extract the object information. In the example, these similarity scores are estimated as $30.68$ and $23.66$. 
If the similarity value decreases, this is considered a success as in the example and we report the percentage of it. 

\begin{table}[]
\centering
\caption{Quantitative results of our method and competing methods on \dataset~dataset. }
\resizebox{1.0\linewidth}{!}{
\begin{tabular}{l|c|c|c|c}
\toprule 
\multirow{2}{*}{Methods} &  \multirow{2}{*}{FID $\downarrow$} & CLIP  & CLIP &  CLIP $\uparrow $ \\
& &Dist. $\uparrow $ & Acc. $\uparrow $  &  Acc. (top 5) \\
\midrule
X-Decoder \cite{zou2022generalized} & 6.360 & 72.2 & 62.6 & 41.5 \\
InstPix2Pix \cite{brooks2022instructpix2pix} & 9.972 & 56.8 & 33.5 & 11.8 \\
CLIPSeg \cite{luddecke2022image} & 8.048 & 71.7 & 57.4 & 33.5 \\
\name~(Ours)  &  \textbf{5.679} & \textbf{76.0} & \textbf{77.4} & \textbf{57.3} \\
\bottomrule
\end{tabular}}
\label{table:results_gqa}
\end{table}

\vspace{0.1cm}\noindent\textbf{CLIP Accuracy.} For this evaluation measure, we utilize CLIP as a zero-shot classifier. In particular, we find out the most likely semantic label of the image region extracted from the source image using the bounding box of the target object, by considering the object classes that exist in our dataset. Next, we perform another prediction this time for the image region extracted from the inpainted image. We present the Top5 predictions as an example in Fig. \ref{fig:clip_score}. We expect the class prediction to change after performing the object removal operation as suggested by the instruction. We mainly look at the Top1 and Top5 predictions. If the predicted class based on the source image is not in the Top1 and Top5 predictions of the inpainted image, that scenario is considered a success.
We report the percentage of successes.

\vspace{0.1cm}\noindent\textbf{RelSim.} We additionally use the relational similarity score proposed by GeNeVa \cite{el2019tell} which is commonly used on CLEVR dataset.
This metric compares the arrangement of objects qualitatively via an object detector and localizer.
We use the object detector released by GeNeVa fine-tuned on CLEVR dataset.
The detector constructs the scene graphs of inpainted and ground-truth images.
The metric computes how many of the ground-truth relations are present in the inpainted image.
We use this score for our evaluations on CLEVR following the previous works \cite{zhang2021text}.
However, we find that this evaluation method is difficult to be used on real images because it requires highly effective object detectors.
For our experiments on \dataset, we use the CLIP-based measures explained below.

\newcommand{\interpfigl}[1]{\includegraphics[trim=0 0 0cm 0, clip, width=2.4cm]{#1}}

\begin{figure*}[!t]
\centering
\scalebox{0.73}{
\addtolength{\tabcolsep}{-5pt}   
\begin{tabular}{cccccccccc}
\\ 
Source & Attn. Map & {Output} & {Attn. Map} & {Output} & Source & Attn. Map & {Output} & {Attn. Map} & {Output} \\
\interpfigl{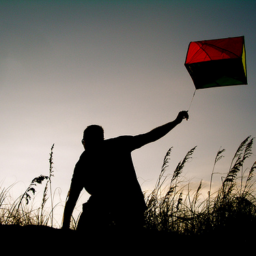} &
\interpfigl{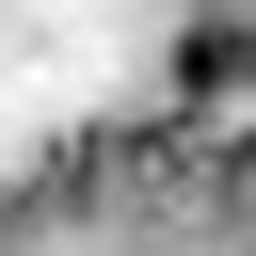} &
\interpfigl{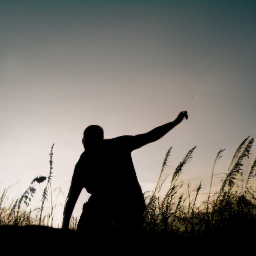} &
\interpfigl{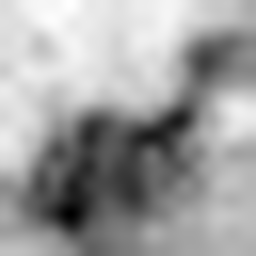} &
\interpfigl{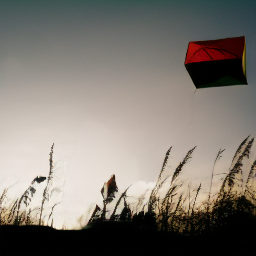}&
\interpfigl{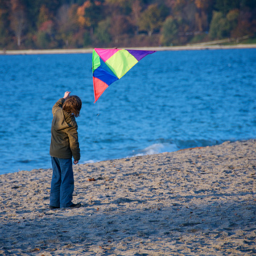} &
\interpfigl{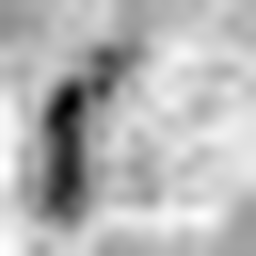} &
\interpfigl{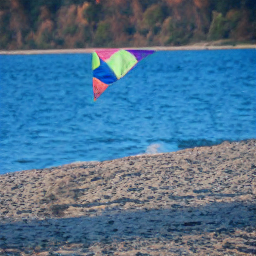} &
\interpfigl{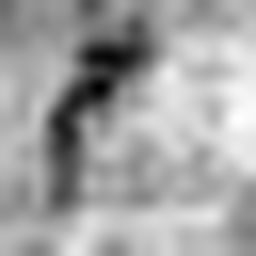} &
\interpfigl{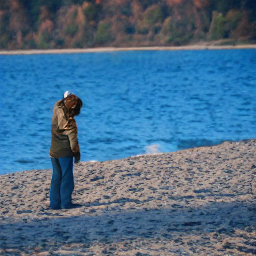}
\\
\ding{229} Remove &
\multicolumn{2}{c}{\{{the flying kite at the right}\} } &
\multicolumn{2}{c}{\{{the man at the center}\}} & 
\ding{229} Remove &
\multicolumn{2}{c}{\{{the boy at the left} \}} &
\multicolumn{2}{c}{\{{the colorful kite at the center} \}}
\\
\interpfigl{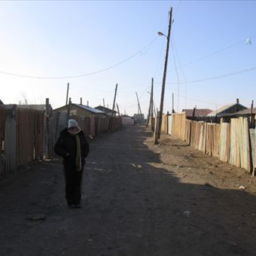} &
\interpfigl{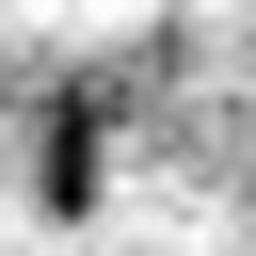} &
\interpfigl{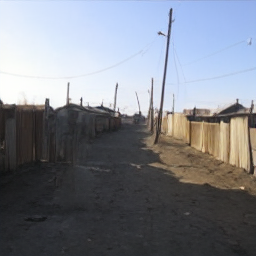} &
\interpfigl{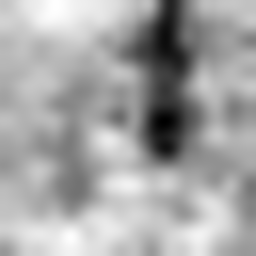} &
\interpfigl{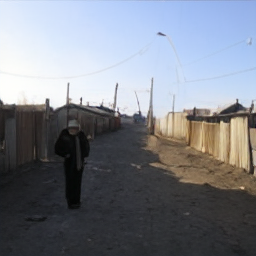} &
\interpfigl{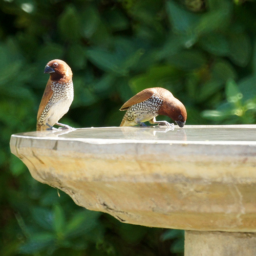} &
\interpfigl{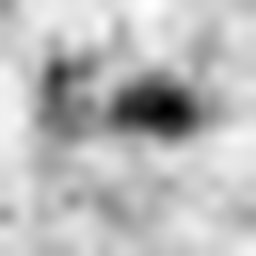} &
\interpfigl{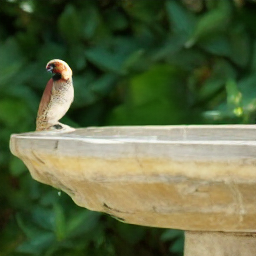} &
\interpfigl{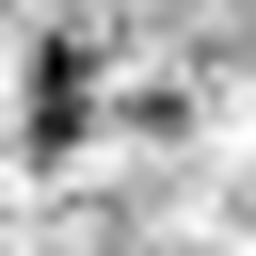} &
\interpfigl{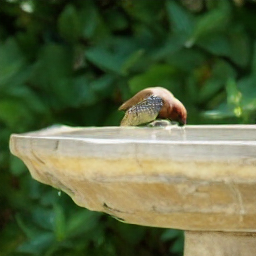}
\\
\ding{229} Remove &
\multicolumn{2}{c}{\{{the person at the center} \}} &
\multicolumn{2}{c}{\{{the pole at the center} \}} & 
\ding{229} Remove &
\multicolumn{2}{c}{\{{the bird at the center}\}} &
\multicolumn{2}{c}{\{{the bird at the left}\}}\\
\interpfigl{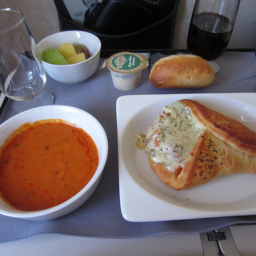} &
\interpfigl{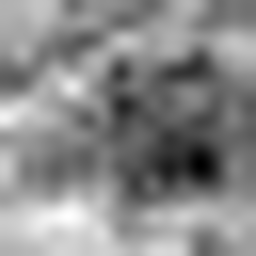} &
\interpfigl{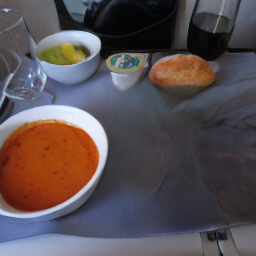} &
\interpfigl{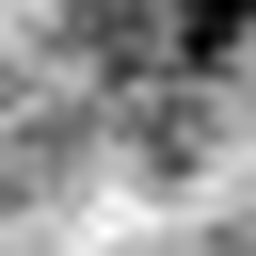} &
\interpfigl{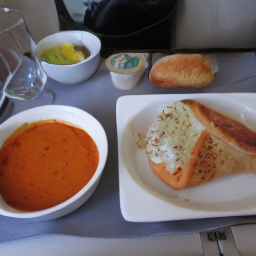} & 
\interpfigl{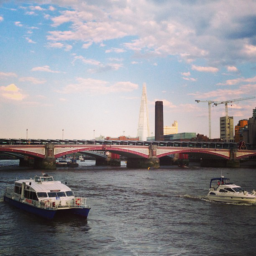} &
\interpfigl{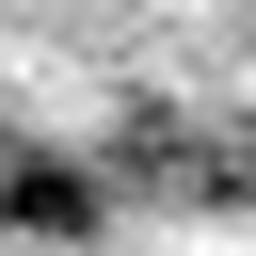} &
\interpfigl{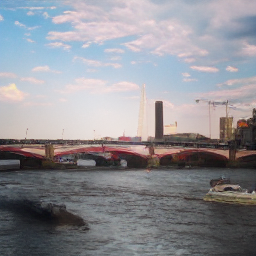} &
\interpfigl{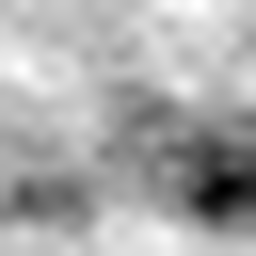} &
\interpfigl{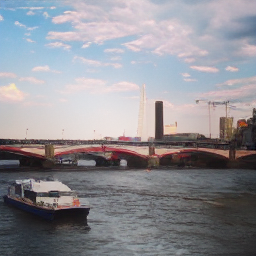}
\\
\ding{229} Remove &
\multicolumn{2}{c}{\{{the square plate on the table}\}} &
\multicolumn{2}{c}{\{{the small glass at the right}\}} &
\ding{229} Remove &
\multicolumn{2}{c}{\{{the boat at the left} \}} &
\multicolumn{2}{c}{\{{the white boat at the right} 
\}}
\\
\interpfigl{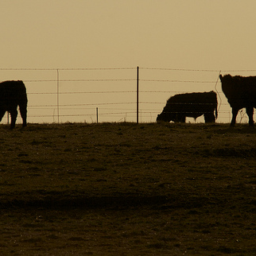} &
\interpfigl{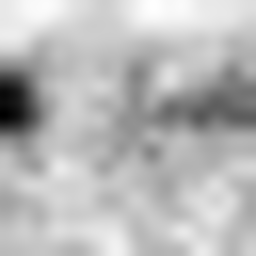} &
\interpfigl{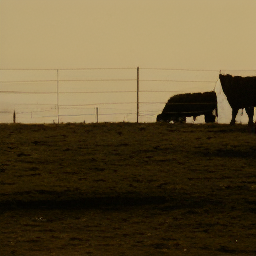} &
\interpfigl{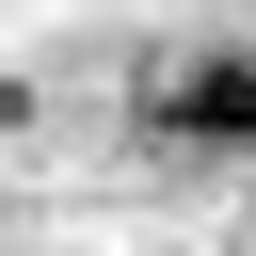} &
\interpfigl{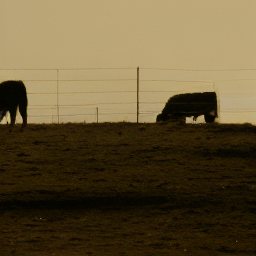} &

\interpfigl{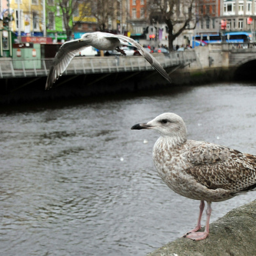} &
\interpfigl{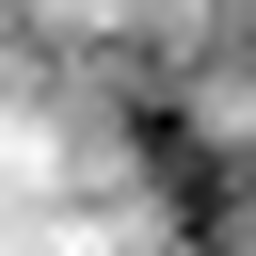} &
\interpfigl{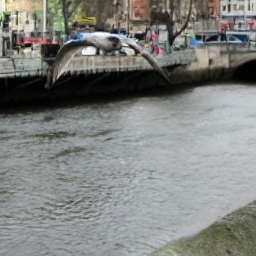} &
\interpfigl{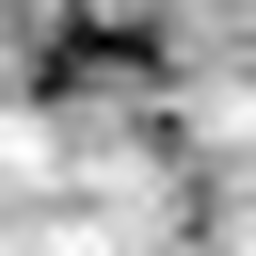} &
\interpfigl{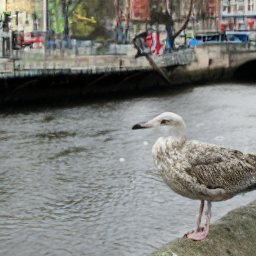}
\\
\ding{229} Remove &
\multicolumn{2}{c}{\{{the cow at the left}\}} &
\multicolumn{2}{c}{\{{the large cow at the right}\}} &
\ding{229} Remove &
\multicolumn{2}{c}{\{{the white bird at the right}\}} &
\multicolumn{2}{c}{\{{the flying bird at the center}\}}
\\
\end{tabular}
}
\caption{Additional instruction-based image inpainting results on the GQA-Inpaint dataset along with the average cross-attention maps of our method.
For the same input image, we show two different inpainting results and their attention maps based on two different instructions.}
\label{fig:attention}
\end{figure*}

\subsection{Experimental Analysis}

\vspace{0.1cm}\noindent\textbf{Results on GQA-Inpaint Dataset.}
We provide the quantitative and qualitative results on \dataset~dataset in Table \ref{table:results_gqa} and Fig. \ref{fig:results_real}, respectively.
We achieve better FID and CLIP scores than X-Decoder, InstPix2Pix, and CLIPSeg as given in Table \ref{table:results_gqa}.
As seen in Fig. \ref{fig:results_real}, InstPix2Pix struggles to achieve the inpainting in many examples. In the first one, instead of removing the gray kite, the model makes the image more gray.
In some examples, it successfully erases the objects and in some, it does not remove the objects at all or remove them partially. Instruct X-decoder struggles to find the correct object to remove and does not perform any editing in many examples such as the first row-second example. 
In the first example, it correctly finds the object to be removed but generates artifacts in the final inpainted image.
Moreover, it sometimes does output interesting results. For example, in many examples, the removed area is filled with an object from the category such as removing the television, boy, and man examples even though the model is instructed to fill the inpainted area in regard to the background. 
ClipSeg also performs and fails similarly to X-Decoder. 
That behavior leads to poor performance for Instruct X-Decoder and ClipSeg in terms of CLIP Score.
When an object is not edited or replaced with the same category, the CLIP score is negatively impacted.
On the other hand, when no edit is applied to the image as in many examples of X-Decoder and ClipSeg, that brings an advantage for the FID score. 
However, our method still achieves a better FID score than other methods and that is because the inpaintings do not result in artifacts, unlike the competing methods. 
Our \name~model successfully erases objects in challenging scenarios.



\vspace{0.1cm}\noindent\textbf{Analysis.}  We visualize the average cross-attention masks in Fig. \ref{fig:attention} for samples of GQA-Inpaint images.
We plot the pixels the prompt attends to.
Because ours is a removal task, we find the attention masks to focus on the objects that will be erased. 
As shown in Fig. \ref{fig:attention}, for the same source image, our model can erase different objects based on the prompts and we see that attention maps focus on those objects that we want to remove.

Next, we analyze the accuracy of attention maps in terms of their segmentation of the prompted object. In our method, we do not explicitly predict masks but inpaint objects in an end-to-end framework.
However, to erase an object, the network needs to detect the pixels to be erased implicitly and we see in Fig. \ref{fig:attention} that attention maps detect where the objects that we want to erase are.
We would like to compare the accuracy of these detections with other methods that predict masks based on prompts such as CLIPSeg \cite{luddecke2022image} and X-Decoder \cite{zou2022generalized}. 
To analyze that, we train a simple UNet that takes attention maps and predicts a mask. 
The UNet only takes the attention maps, it does not take prompts or image features as inputs. 
This UNet is trained separately with the sole purpose of analyzing the mask predictions of a pretrained Inst-Inpaint.
The attention maps are resized to $32\times32$ resolution from 16x16, 8x8, and 4x4 resolutions by using nearest-neighbor interpolation. 
We simply concatenate attention maps from all time-frames which brings a channel of $1280$ inputs.
The network has 4 downsampling and 4 upsampling residual block layers with channels going as $\{32, 64, 96, 128, 128, 96, 64, 32\}$. In the end, we have a bilinear upsampling layer which brings the resolution to $256\times256$. This is a very simple set-up as we expect the attention maps to already generate well-defined predictions.
We compare the mask accuracies in Table \ref{table:results_iou}.
Our method achieves better scores than ClipSeg and X-Decoder.

We visualize the mask predictions in Fig. \ref{fig:results_iou} of X-Decoder, ClipSeg, and ours that come from the UNet model which takes the attention maps as inputs.
Our model is better at predicting the correct masks as was also expected based on the inpainting results.
Note that, we set the U-Net to work on $32\times32$ resolution and only add a bilinear upsampling in the end, better scores with higher details on the mask can be obtained with a more powerful decoder.
We set the segmentation network simple for our analysis.

\begin{table}[]
\centering
\caption{Segmentation scores (IoU) of mask predictions on \dataset~datasets. 
Our method does not explicitly detect a semantic map. We train a simple network that operates on our attention maps to evaluate the accuracy of detections.}
\resizebox{0.8\linewidth}{!}{
\begin{tabular}{c|c|c}
\toprule 
CLIPSeg \cite{luddecke2022image} & X-Decoder \cite{zou2022generalized} & Inst-Inpaint (Ours) \\
\toprule
40.6 & 48.7 & 53.1 \\
\bottomrule
\end{tabular}
}
\label{table:results_iou}
\end{table}

\newcommand{\interpfigt}[1]{\includegraphics[trim=0 0 0cm 0, clip, width=1.8cm]{#1}}
\newcommand{\interpfigtprompt}[1]{\includegraphics[trim=0 -1cm 0 0, clip, width=2.0cm]{#1}}

\begin{figure}[]
\centering
\scalebox{0.9}{
\addtolength{\tabcolsep}{-5pt}   
\begin{tabular}{cccccc}
\\ 
Source &  X-Decoder & ClipSeg & Ours & GT \\
\interpfigt{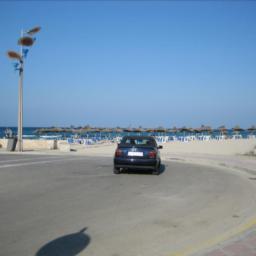} &
\interpfigt{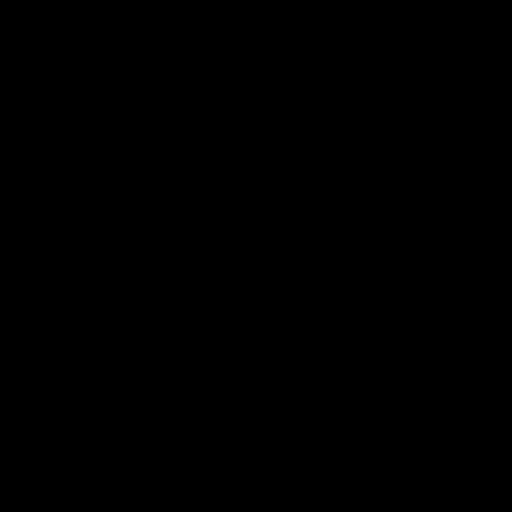} &
\interpfigt{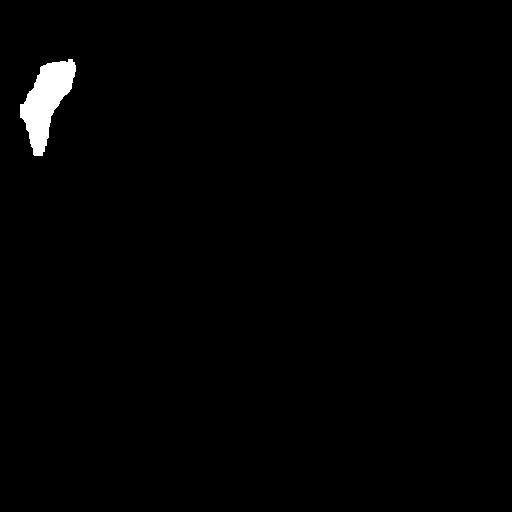} &
\interpfigt{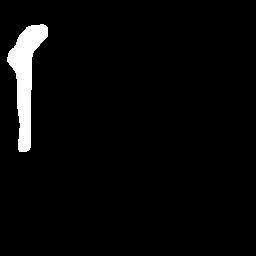} &
\interpfigt{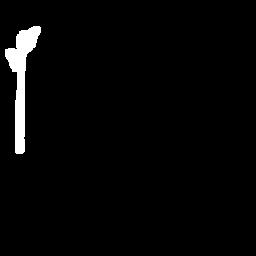}
\\
\ding{229} Remove &
\multicolumn{3}{l}{the street light at the left} 
\\
\interpfigt{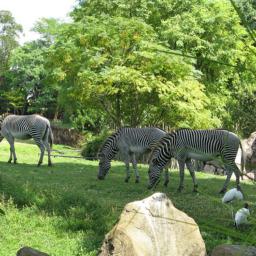} &
\interpfigt{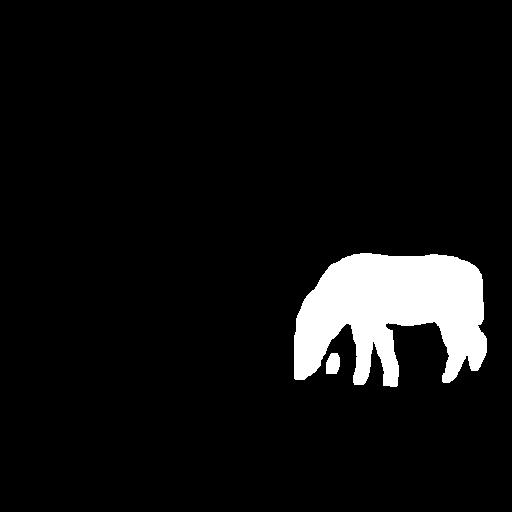} &
\interpfigt{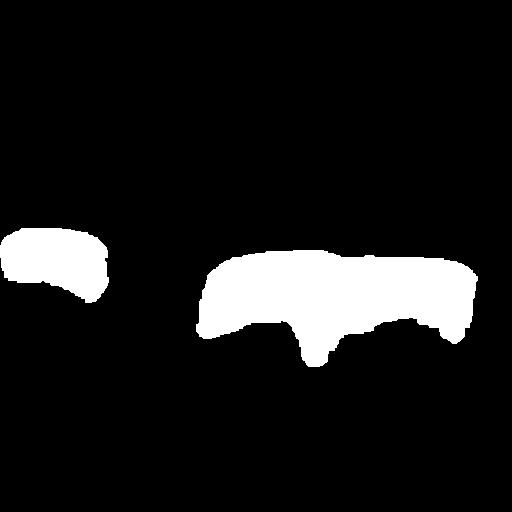} &
\interpfigt{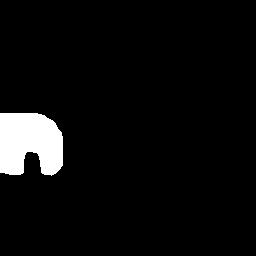} &
\interpfigt{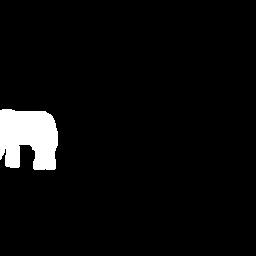}
\\
\ding{229} Remove &
\multicolumn{3}{l}{the standing zebra at the left} 
\\
\interpfigt{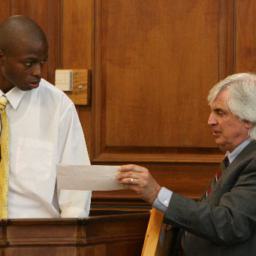} &
\interpfigt{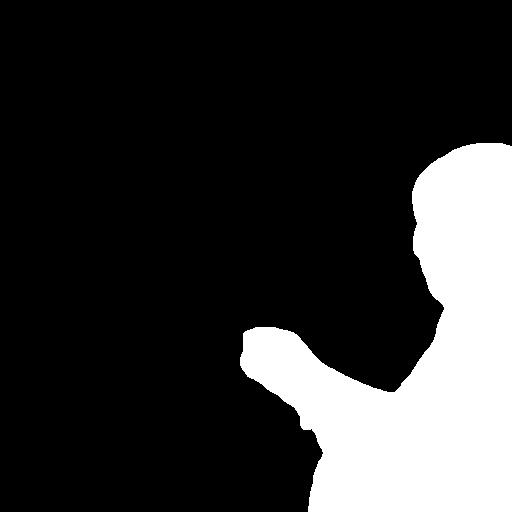} &
\interpfigt{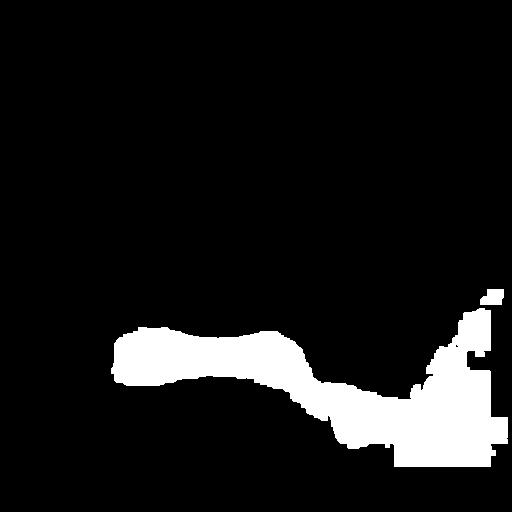} &
\interpfigt{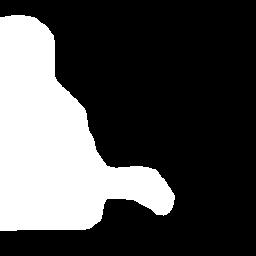} &
\interpfigt{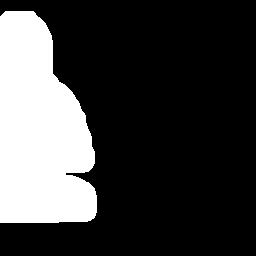}
\\
\ding{229} Remove &
\multicolumn{3}{l}{the man at the left of the paper} 
\\
\interpfigt{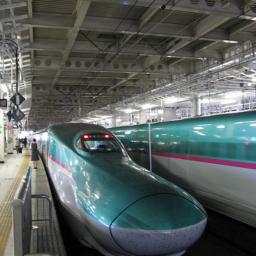} &
\interpfigt{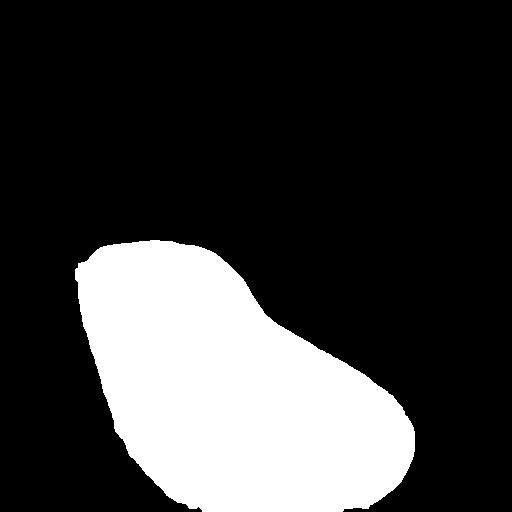} &
\interpfigt{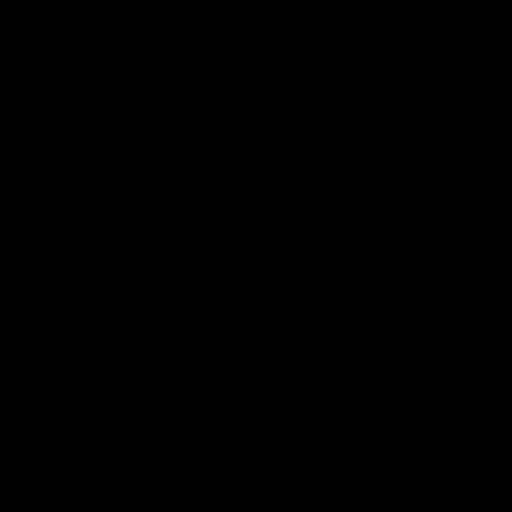} &
\interpfigt{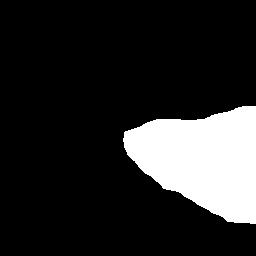} &
\interpfigt{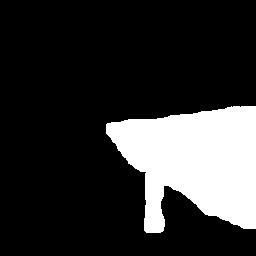}
\\
\ding{229} Remove &
\multicolumn{3}{l}{the green train at the right} 
\\
\end{tabular}
}
\caption{Qualitative results of our segmentation analysis and the mask predictions of the competing methods on the GQA-Inpaint dataset.}
\label{fig:results_iou}
\end{figure}

\begin{figure}[]
\centering
\scalebox{0.9}{
\addtolength{\tabcolsep}{-5pt}   
\begin{tabular}{cccccc}
\\ 
Source &  cManiGAN & TIM-GAN & Ours & GT \\

\interpfigt{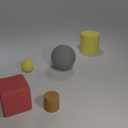} &
\interpfigt{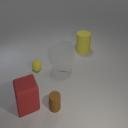} &
\interpfigt{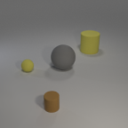} &
\interpfigt{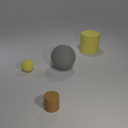} &
\interpfigt{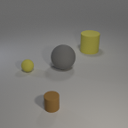}
\\
\ding{229} Remove &
\multicolumn{3}{l}{the red object} \\
\interpfigt{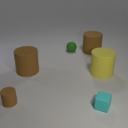} &
\interpfigt{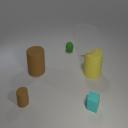} &
\interpfigt{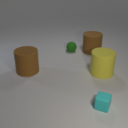} &
\interpfigt{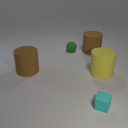} &
\interpfigt{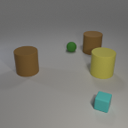}
\\
\ding{229} Remove &
\multicolumn{3}{l}{bottom-left small brown object} \\
\interpfigt{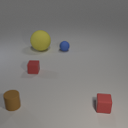} &
\interpfigt{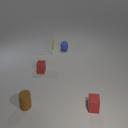} &
\interpfigt{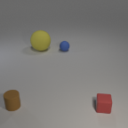} &
\interpfigt{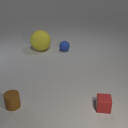} &
\interpfigt{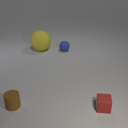}
\\
\ding{229} Remove &
\multicolumn{3}{l}{middle-left object} \\
\interpfigt{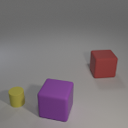} &
\interpfigt{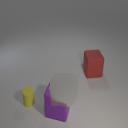} &
\interpfigt{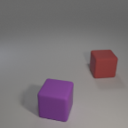} &
\interpfigt{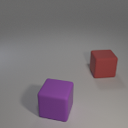} &
\interpfigt{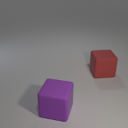}
\\
\ding{229} Remove &
\multicolumn{3}{l}{bottom-left small object} \\
\interpfigt{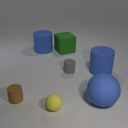} &
\interpfigt{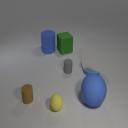} &
\interpfigt{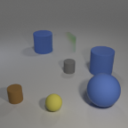} &
\interpfigt{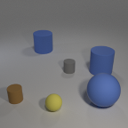} &
\interpfigt{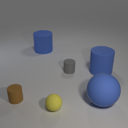} \\
\ding{229} Remove &
\multicolumn{3}{l}{top-center large object} \\
\end{tabular}
}
\caption{Qualitative results of our approach and the competing methods on synthetic CLEVR dataset.}
\label{fig:results_clevr}
\end{figure}

\vspace{0.1cm}\noindent\textbf{Comparisons with GAN-based models on CLEVR Dataset.} Next, we run experiments with the more simple CLEVR dataset and compare our method with GAN-based methods on this dataset.
We provide the quantitative and qualitative results on the CLEVR dataset in Table \ref{table:results_clevr} and Fig. \ref{fig:results_clevr}, respectively.
We train TIM-GAN with their released implementation. 
TIM-GAN, in the author's paper, is extensively tested on the CLEVR dataset and there are provided pre-trained models. 
However, we train their model again since we are only interested in the remove instruction.
For cMANIGAN, the authors provide us with the output images based on the text and input images we send them.

cManiGAN is trained on the CLEVR dataset but not by us. The authors mention that the caption templates they use differ from ours and might result in poor performance. 
The visual and quantitative results of cManiGAN are indeed poor in our samples. 
The object to be removed is not correctly detected in many examples and furthermore, the objects are not inpainted with high-fidelity.
There is a ghosting effect in the first example and objects are partially removed in some examples (last 4 examples).
We achieve a better FID score and a comparable RelSim score to those of TIM-GAN. Visually, inpainting results obtained by TIM-GAN include artifacts as it is not always able to fully remove the objects. Our diffusion based \name~model, on the other hand, does include less artifacts on this simple dataset. 
Note that this is a quite easy set-up compared to the GQA-Inpaint dataset.
We share these results to compare with GAN-based models that are built on this dataset and to show that \name~model set-up  achieves better results than the proposed GAN-based models.
Because other diffusion-based models, X-Decoder, CLIPSeg, and InstPix2Pix are trained on a large number of real images, we do not include them in this comparison.

\begin{table}[]
\centering
\caption{Quantitative results of our method and competing methods on synthetic CLEVR dataset for the instruction-based inpainting task.}
\resizebox{0.7\linewidth}{!}{
\begin{tabular}{l|c|c}
\toprule 
Methods&  FID $\downarrow $ & RelSim $\uparrow $ \\
\midrule
cManiGAN \cite{fan2022target} & 27.61 & 0.613 \\
TIM-GAN \cite{zhang2021text} & 0.558 & 0.919 \\
\name~(Ours)  & \textbf{0.453} & \textbf{0.914}  \\
\bottomrule
\end{tabular}}
\label{table:results_clevr}
\end{table}

\subsection{Model Limitations}
\label{sec:limitations}

We train our model by taking LDM as the base architecture, which applies diffusion steps in the latent space instead of the pixel space itself. For this purpose, the model uses an autoencoder model to obtain the latent representations of the images, which is trained in the first stage. For simple datasets, such as CLEVR, the reconstructed images are identical to the source images. However, in complex datasets such as GQA-Inpaint, some scenes with complex patterns especially texts can be  poorly reconstructed by the autoencoder. Even if our model successfully finds and removes the target object based on the given instruction, the output quality can be poor due to this reason. 
An example can be seen in Figure~\ref{fig:teaser} - remove the colorful train, the writing on the input image is not reconstructed successfully in the output.
More powerful autoencoders or different optimization methods can be built to prevent these problems. 
We hope the release of the dataset, baseline models, and our analysis will facilitate research in this domain.

\section{Conclusion}
In this work, we introduced a novel instruction-based real image inpainting task, where the unwanted objects are specified only via textual instructions without the need for binary masks. Accordingly, we developed a diffusion-based framework, \name, that can remove objects from images based solely on textual prompts. We also created a new benchmark dataset, \dataset, and suggested novel evaluation measures for this task. Our experimental results show that our framework outperforms the existing methods in this task. Our work has demonstrated the potential of using textual instructions as a more user-friendly and natural way of controlling image inpainting. We will release both our constructed dataset, \dataset, and our model, \name,  to facilitate research in this domain.

\ifCLASSOPTIONcaptionsoff
  \newpage
\fi

%

{\small
\bibliographystyle{ieee}
\bibliography{references}
}

\end{document}